%% file: main.tex
\PassOptionsToPackage{table,dvipsnames}{xcolor}

\documentclass[10pt,twocolumn,letterpaper]{article}

\usepackage[pagenumbers]{iccv} 

\usepackage{graphicx}
\usepackage{booktabs}
\usepackage{caption}
\usepackage{footmisc}
\usepackage{multirow}
\usepackage{adjustbox}
\usepackage{subcaption}
\usepackage{amsmath}
\usepackage{wasysym}
\usepackage{booktabs}
\usepackage{algorithm}
\usepackage{algpseudocode}
\usepackage{amssymb}
\usepackage{utfsym}
\usepackage[table]{xcolor} 
\usepackage{listings}
\usepackage{footmisc}
\renewcommand\thefootnote{\arabic{footnote}}

\definecolor{uclablue}{RGB}{159, 195, 224}

\definecolor{uclagold}{RGB}{254,180,167}

\definecolor{grayred}{RGB}{232,237,205}

\newcommand{\dataset}{GenRef\xspace}
\newcommand{\model}{ReflectionFlow\xspace}

\newif\ifpaper

\definecolor{iccvblue}{rgb}{0.21,0.49,0.74}
\usepackage[pagebackref,breaklinks,colorlinks,allcolors=iccvblue]{hyperref}

\usepackage[capitalize]{cleveref}
\usepackage{manyfoot}%
\usepackage{booktabs}%
\usepackage{makecell} 
\usepackage{multirow}

\def\paperID{0000} 
\def\confName{ICCV}
\def\confYear{2025}

\title{From Reflection to Perfection: Scaling Inference-Time Optimization for Text-to-Image Diffusion Models via Reflection Tuning}

\author{
    Le Zhuo\textsuperscript{\rm 1, 4}\footnotemark[1],
   Liangbing Zhao\textsuperscript{\rm 2}\footnotemark[1], Sayak Paul\textsuperscript{\rm 3}, Yue Liao\textsuperscript{\rm 1}, Renrui Zhang\textsuperscript{\rm 1},
  Yi Xin\textsuperscript{\rm 4} \\
  Peng Gao\textsuperscript{\rm 4}, Mohamed Elhoseiny\textsuperscript{\rm 2}\footnotemark[2], Hongsheng Li\textsuperscript{\rm 1}\footnotemark[2]\\
  \textsuperscript{\rm 1}{CUHK MMLab}, \textsuperscript{\rm 2}{KAUST},
  \textsuperscript{\rm 3}{Hugging Face}, 
  \textsuperscript{\rm 4}{Shanghai AI Lab}\\
  \small Project page:~\url{https://diffusion-cot.github.io/reflection2perfection}
}

\begin{document}

\twocolumn[{
\renewcommand\twocolumn[1][]{#1}
\maketitle
\begin{center}
    \captionsetup{type=figure}
    \centering
    \includegraphics[width=\textwidth]{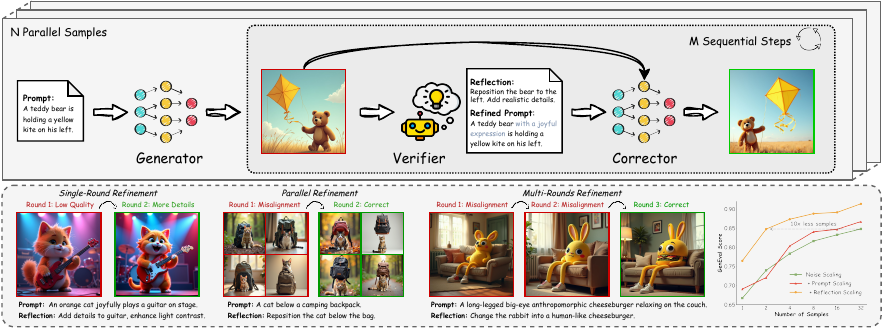}
    \caption{Overall pipeline of the \model framework with qualitative and quantitative results of scaling compute at inference time.}
        \label{fig:overview}
\end{center}
}]

\maketitle

\let\tempfootnote\thefootnote
\let\thefootnote\relax
\footnotetext{$^*$ Equal Contribution}
\footnotetext{$\dagger$ Corresponding Author}

\let\thefootnote\tempfootnote
\renewcommand\thefootnote{\arabic{footnote}}

\input{sec/0_abstract}    
\input{sec/1_intro}
\input{sec/2_related}
\input{sec/3_method}

\input{sec/4_exp}
\input{sec/5_conclusion}

{
    \small
    \bibliographystyle{ieeenat_fullname}
    \bibliography{main}
}
\ifpaper
\else
    \clearpage
    \newpage
    \onecolumn
    \appendix
    \setcounter{section}{0}
    \def\thesection{\Alph{section}}
    \section*{Appendix}
    \input{sec/6_appendix}

\fi 

\end{document}

%% file: sec/0_abstract.tex
\begin{abstract}
Recent text-to-image diffusion models achieve impressive visual quality through extensive scaling of training data and model parameters, yet they often struggle with complex scenes and fine-grained details. Inspired by the self-reflection capabilities emergent in large language models, we propose \model, an inference-time framework enabling diffusion models to iteratively reflect upon and refine their outputs. \model introduces three complementary inference-time scaling axes: (1) noise-level scaling to optimize latent initialization; (2) prompt-level scaling for precise semantic guidance; and most notably, (3) reflection-level scaling, which explicitly provides actionable reflections to iteratively assess and correct previous generations. To facilitate reflection-level scaling, we construct \dataset, a large-scale dataset comprising 1 million triplets, each containing a reflection, a flawed image, and an enhanced image. Leveraging this dataset, we efficiently perform reflection tuning on state-of-the-art diffusion transformer, FLUX.1-dev, by jointly modeling multimodal inputs within a unified framework. Experimental results show that \model significantly outperforms naive noise-level scaling methods, offering a scalable and compute-efficient solution toward higher-quality image synthesis on challenging tasks. All code, checkpoints, and datasets are available at \url{https://diffusion-cot.github.io/reflection2perfection}.
\end{abstract}

%% file: sec/1_intro.tex
\section{Introduction}
\label{sec:intro}

Recent advances in text-to-image~(T2I) diffusion models~\cite{esser2024scaling,flux2024,zhuolumina,xie2024sana,guo2025can, abdelrahman2023toddlerdiffusion,lumina2} have led to remarkable progress in image synthesis. 
With the scaling of training resources, their capability to create high-resolution and photorealistic images steadily improves. Despite their success, their performance across various domains remains inconsistent, especially when tasked with generating complex human poses, multiple-object compositions, or scenes with complicated lighting and shadows. To alleviate these limitations, it is necessary to exponentially scale the training compute, model parameters, and data size, according to established scaling laws at training time~\cite{kaplan2020scaling}.

This motivates us to rethink the prevailing paradigm:~\emph{instead of continuously scaling pretrained models, how to effectively exploit the full capabilities of existing diffusion models during inference?} 
We hypothesize that while T2I models may struggle to generate desirable images within a fixed compute budget, their performance can be significantly improved through an iterative refinement process. This idea of leveraging additional computational resources to enhance performance during inference has been validated in large language models (LLMs)~\cite{snell2024scaling}. Modern LLMs have demonstrated the ability to improve their outputs by reflecting on intermediate outputs and subsequently refining their responses~\cite{madaan2023self,kumar2024training,shinn2023reflexion}, leading to superior performance on complex tasks such as math problem solving and code generation. However, current inference-time optimizations for diffusion models~\cite{ma2025inference,singhal2025general,xie2025sana} focus on parallelized noise-space search strategies, leaving sequential self-refinement paradigm that could enable diffusion models to reflect upon and correct their own mistakes largely unexplored.

In this paper, we introduce \textbf{\model}, a novel inference-time self-refinement framework for diffusion models leveraging iterative reflection. Our approach explores three dimensions for scaling inference-time computation: (1) noise-level scaling, which searches for better noise initialization; (2) prompt-level scaling, which optimizes input prompt for precise semantic guidance; and (3) reflection-level scaling, which constructs explicit reflections to iteratively assess and correct previously generated outputs. Integrating these dimensions enables diffusion models to flexibly exploit additional computational resources at inference-time, continuously improving the image quality through an iterative process as illustrated in~\cref{fig:overview}. 

%

A fundamental challenge underlying our approach is: \emph{how to empower T2I diffusion models with self-refinement capabilities?}
Currently, no existing diffusion model is able to accurately interpret reflection prompts and leverage previously generated images for iterative refinement. Reflecting on the intrinsic similarity between self-refinement and image editing tasks, we observe that both involve generating a new output, jointly conditioned on textual and visual contexts. Inspired by the training paradigm of efficiently adapting diffusion priors to image 
editing~\cite{brooks2023instructpix2pix,wei2024omniedit}, we hypothesize that adopting similar methods could enable diffusion models to achieve self-refinement. However, another critical challenge arises: the absence of dedicated datasets specifically designed for reflection-guided refinement.

Motivated by this insight, we propose \textbf{\dataset}, the first large-scale image reflection dataset comprising 1 million reflection triplets in multiple domains. To ensure the validity of these triplets (\emph{e.g.}, flawed images accurately capture potential errors, the reflections offer effective and actionable editing suggestions, and the good images exhibit sufficiently higher quality), we design a scalable, fully automatic data construction pipeline with four distinct data sources, leveraging verifiable objectives, ensemble reward models, and diverse rollout strategies. We further create a chain-of-thought (CoT) reflection subset, named \textbf{\dataset-CoT}, providing 227K high-quality progressive annotations from GPT-4o~\cite{hurst2024gpt} and Gemini 2.0~\cite{team2023gemini}. We then employ an efficient reflection tuning tailored for T2I diffusion transformers, such as FLUX.1-dev~\cite{flux2024}. In our approach, the original prompt, reflection prompt, flawed image, and high-quality image are concatenated into a single unified sequence to perform joint multimodal attention, thereby eliminating the need for additional modules~\cite{zhang2023adding, ye2023ip}. Note that beyond laying the foundation for reflection tuning, our carefully curated dataset can also serve broader applications, such as general preference tuning~\cite{wallace2024diffusion} and reward modeling~\cite{liu2025improving}.

Through experiments, \model demonstrates the potential for enhancing image generation quality without requiring additional large-scale training. Compared to naive noise-level scaling, \model achieves significantly better performance under identical inference budgets, with performance consistently improved using better verifiers. Additionally, we explore the flexible trade-off between search width and reflection depth, as well as overall inference budgets. Further evaluation shows that our approach achieves particularly substantial gains on challenging prompts. Qualitative analyses illustrate how our model iteratively reflects on and corrects its outputs, progressively converging to superior solutions.

In summary, \textbf{(1)} we propose \model, a reflection-based inference-time scaling framework for diffusion models, integrating three scaling axes,~\emph{i.e.}, noise-, prompt-, and reflection-level; \textbf{(2)} we propose a scalable pipeline for collecting high-quality image reflection data and introduce \dataset, the first large-scale dataset containing 1 million triplets, along with an additional 227K CoT reflection annotations; \textbf{(3)} we employ an efficient reflection tuning method for diffusion transformers with a tailored training strategy; \textbf{(4)} extensive experiments demonstrate that \model substantially boosts performance via scaling inference-time compute, unlocking reasoning capability.

%% file: sec/2_related.tex
\section{Related Work}
\label{sec:related}

\noindent\textbf{Text-to-Image Diffusion Models.} 
Text-to-image diffusion models have rapidly advanced in terms of model architecture and training strategies. In terms of architecture evolution, the community has transitioned from the prevalent U-Net diffusion models~\cite{rombach2022high} towards diffusion transformers (DiT)~\cite{dit}. Regarding training strategies, earlier complex hand-designed diffusion schedules~\cite{ho2020denoising,songscore} have given way to simpler flow-based formulations~\cite{liu2022flow,lipman2022flow}, significantly enhancing training efficiency through techniques such as multi-resolution progressive training~\cite{chen2024pixart}. Recently, scaling up both the training datasets and model parameters has led to the emergence of various large-scale, flow-based diffusion transformer models~\cite{esser2024scaling,flux2024,gaolumina,zhuolumina,xie2024sana,qin2025lumina}.

\noindent\textbf{Diffusion Inference-Time Enhancement.} 
Building upon the powerful pretrained diffusion models, recent research has increasingly focused on unleashing their potential at inference time. One line of research has observed that the initial noise significantly impacts generation quality~\cite{zhou2024golden,qi2024not}, prompting methods to identify superior initialization strategies~\cite{ma2025inference,zhou2024golden,ahn2024noise,qi2024not}. Another research direction aims to improve the iterative sampling procedure of diffusion models~\cite{ye2024tfg,bai2024zigzag,singhal2025general}, notably through denoising and inversion~\cite{songdenoising}. Additionally, recent studies~\cite{li2024hunyuan,xie2024sana} demonstrate that augmenting input prompts can substantially improve visual fidelity and text-image alignment. While existing methods focus on parallel single-pass generation, our work proposes a sequential generation then refinement procedure, integrating both parallel and sequential inference-time scaling into a unified framework.

\noindent\textbf{Scaling Inference-Time Compute.} 
Recent studies on LLMs have provided valuable insights into inference-time scaling laws. A primary line of investigation~\cite{feng2023alphazero,yao2023tree} has explored search algorithms, such as best-of-N and beam search, with verifiers to select higher-quality outputs. Another prominent direction focuses on enabling LLMs to refine their own outputs. For instance, techniques~\cite{kim2023language,madaan2023self,shinn2023reflexion} such as zero-shot prompting have been employed to elicit self-reflection from models, enabling them to iteratively enhance their outputs. Furthermore, supervised fine-tuning (SFT) and reinforcement learning (RL) approaches~\cite{deepseekr1,welleckgenerating,kumar2024training,qu2024recursive,havrillaglore} have also been introduced to explicitly train models for reflective self-improvement. 
One recent work~\cite{guo2025can} investigated RL, test-time scaling, and reflection on autoregressive image generation models, providing preliminary insights into this field. 
Meta also presented a comprehensive framework~\cite{snell2024scaling} that systematically unifies these directions, which thoroughly investigates the trade-offs between the pretraining scale and the computation of inference time, significantly inspiring our approach.

%% file: sec/3_method.tex
\section{Method}
\label{sec:method}

\subsection{Problem Formulation}
\label{sec:setup}
We first formalize the iterative refinement framework for the text-to-image (T2I) generation task. Given a textual prompt $y$ and a T2I generator $G_{\theta}$, we initially generate an image $x_0 \sim G_{\theta}(\cdot \mid z, y)$ from random noise $z \in \mathcal{N}(0,I)$. Subsequently, during each refinement iteration $i$, we introduce a corrector model $C_{\phi}$ to produce an improved image $x_i \sim C_{\phi}(\cdot \mid z, y, x_{i-1}, r_{i-1})$, conditioned on the previous iteration's image $x_{i-1}$, the original textual prompt $y$, and a textual reflection $r_{i-1}$ prompt describing previous shortcomings and improvement directions. Specifically, we utilize external evaluation modules such as reward models~\cite{hessel-etal-2021-clipscore,NEURIPS2023_33646ef0} and multimodal large language models (MLLMs)~\cite{hurst2024gpt,Qwen2.5-VL} to assess the quality of generated image at iteration $i-1$ and produce an instructive reflection $r_{i-1} \sim R(\cdot \mid x_{i-1}, y)$. Through this iterative refinement process, the original T2I generation task is reformulated into a sequential generation-and-refinement paradigm, expressed mathematically as:

\begin{equation}
    G_{\theta}(x_N \mid z, y) = G_{\theta}(x_0 \mid z, y)\prod_{i=1}^{N} C_{\phi}(x_i \mid z, y, x_{i-1}, r_{i-1}).
    \label{eq:refine}
\end{equation}

A central challenge in this iterative refinement approach lies in effectively training the corrector model $C_{\phi}$ to recognize and rectify its own errors based on textual reflections. Inspired by recent advances~\cite{brooks2023instructpix2pix,wei2024omniedit} in image editing, we observe that the objective of our corrector $C_\phi$ defined in Equation~\ref{eq:refine} closely resembles the general image editing problem, \emph{i.e.}, transforming an input image into a desired target based on textual instructions. Thus, we conceptualize the self-correction task, guided by textual reflections, as a generalized editing problem. By constructing a dedicated reflection dataset (Section~\ref{sec:dataset}) and performing efficient reflection tuning (Section~\ref{sec:training}), we enable a foundational T2I diffusion model to effectively learn to refine and iteratively improve its own generations (Section~\ref{sec:infer}).

\subsection{Reflection Dataset}
\label{sec:dataset}

\begin{figure*}[t]
    \centering
    \includegraphics[width=\textwidth]{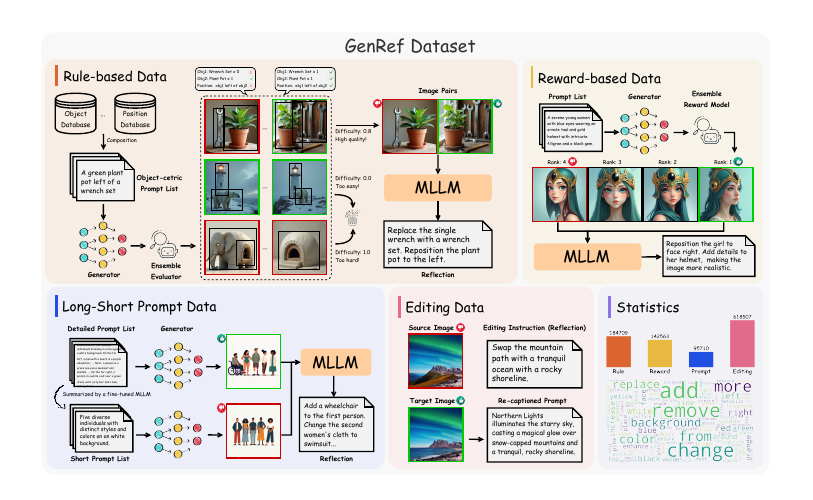}
    \caption{Construction pipelines and statistics of our \dataset dataset. We collect our reflection triplets (flawed images, enhanced images, textual reflections) from four distinct data sources, including: rule-based data, reward-based data, long-short prompt data, and editing data.}
    \label{fig:data}
\end{figure*}

The key to enabling a T2I diffusion model to learn self-refinement lies in constructing an appropriate dataset. However, there is currently a lack of suitable open-source datasets specifically curated for image refinement tasks guided by textual feedback. To bridge this gap, we introduce \textbf{\dataset}, the first large-scale T2I refinement dataset comprising 1 million triplets of the form (flawed image, high-quality image, reflection) collected across multiple domains using our scalable pipeline, as illustrated in \cref{fig:data}. Additionally, we gather 227K chain-of-thought (CoT) image reflections annotated by closed-source APIs~\cite{team2023gemini,hurst2024gpt}, which provide detailed pairwise analyses, preference annotations, and reflections. These annotations form the foundation for training a dedicated MLLM verifier capable of reward modeling and reflection generation, thereby being beneficial beyond the task of T2I self-refinement. We provide samples of our datasets in Appendix~\ref{app:dataset}.

\noindent\textbf{Principles.} Inspired by recent advances in self-refinement dataset construction from the LLMs literature~\cite{fineweb}, we first establish several guiding principles that our dataset must satisfy: (1) the ``flawed images'' should comprehensively cover common errors encountered by the generator $G_{\theta}$ during inference, (2) the ``high-quality images'' should clearly exhibit substantial quality improvements relative to their corresponding ``flawed images'' evaluated in diverse aspects, and (3) the textual reflections should accurately describe the observed shortcomings and provide actionable guidance for refinement. Guided by these principles, we develop an automated and scalable pipeline for constructing the dataset, spanning four distinct domains below.

\noindent\textbf{Rule-based Data.} Recent studies, such as Deepseek-R1~\cite{deepseekr1}, have demonstrated that high-quality, rigorously verified datasets constructed via rule-based verifiers significantly enhance model capabilities. Inspired by these methodologies, we first employ GPT-4o~\cite{hurst2024gpt} to brainstorm a diverse list of common object categories along with their associated attributes, such as colors and spatial positions. We subsequently composite these objects and attributes using rule-based methods to construct unique textual prompts. We rigorously filter these prompts to guarantee there is no overlap with test samples from existing benchmarks~\cite{ghosh2024geneval}.

Next, we utilize FLUX.1-dev~\cite{flux2024} to generate 10 candidate images per prompt. Each candidate is verified by Grounded SAM~\cite{ren2024grounded} to detect and localize individual objects, subsequently evaluating attributes such as color and quantity with respect to the provided prompt. Based on these evaluations, each image is assigned either a binary correctness label or a continuous correctness score ranging between 0 and 1, along with explicit identification of the reasons for any detected errors. Afterward, we estimate the difficulty of each prompt defined by the fraction of correct samples. We then bin the difficulty into three quantiles, where prompts yielding predominantly correct or incorrect images, indicating insufficient or excessive difficulty, are discarded. For the remaining challenging prompts, we constructed image refinement pairs by randomly pairing the highest-scoring images with the lowest-scoring images. 

\noindent\textbf{Reward-based Data.} While the rule-based triplets primarily emphasize object-centric text-image alignment, aesthetic quality, visual fidelity, and alignment with general user prompts are equally essential for a universal image refinement model. To address this, we collect a diverse set of general-purpose prompts, including both image captions and real user-generated prompts. For each prompt, we generate 10 candidate images and assess each generated image based on an ensemble scoring strategy utilizing multiple reward models, \emph{e.g.}, HPSv2 score~\cite{wu2023human}, CLIP score~\cite{hessel-etal-2021-clipscore}, and PickScore~\cite{kirstain2023pickapic}, to comprehensively evaluate their overall quality and alignment with the provided prompts. We construct refinement pairs by randomly pairing images from the top-k and bottom-k subsets for each prompt. 

\noindent\textbf{Long-Short Prompt Data.} Recent studies~\cite{betker2023improving, segalis2023pictureworththousandwords, spright} indicate that, for the same intended content, generated image quality consistently improves as textual prompts become more detailed and descriptive. Motivated by this observation, we fine-tune an MLLM specifically to condense detailed prompts annotated by GPT-4o into shorter, more concise versions. Images generated from these two prompt variants naturally form corresponding image pairs, with the detailed prompt-generated images serving as higher-quality instances. To ensure the quality and effectiveness of these pairs, we also generate multiple samples in parallel and employ the ensemble-reward scoring approach for filtering.

\noindent\textbf{Editing Data.} Finally, we augment our dataset with existing image editing datasets to further enhance its diversity and richness. Image editing data inherently provides paired images accompanied by textual editing instructions. By treating the caption of the edited image as the input prompt, and the source and edited images as flawed and high-quality counterparts respectively, we seamlessly integrate editing data into our refinement dataset paradigm. Although the editing domain differs from our primary image refinement tasks, the precise and explicit nature of editing instructions can effectively enhance the model's ability to follow textual guidance and understand precise correspondences between source and target images. Concretely, we select high-quality editing samples from the OmniEdit dataset~\cite{wei2024omniedit} and generate detailed synthetic captions for each edited image using our in-house captioning model.


 \begin{figure}[t]
    \centering
    \includegraphics[width=1.0\linewidth]{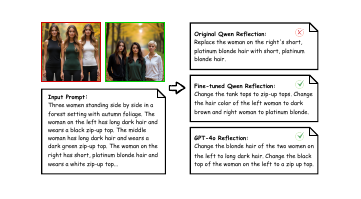}
    \caption{Comparisons of textual reflection generated by original Qwen2.5-VL-7B, our fine-tuned image reflector, and GPT-4o.}
    \label{fig:reflection}
\end{figure}

\noindent\textbf{Reflection Annotation.} After constructing diverse image pairs, it is essential to annotate them with textual reflections that explicitly describe how to transform a flawed image into its corresponding higher-quality counterpart. We experiment with various MLLMs and observed that even the current state-of-the-art open-source model, Qwen2.5-VL~\cite{Qwen2.5-VL}, tends to generate inaccurate reflections when prompted in a zero-shot manner, exhibiting severe hallucinations as illustrated in Fig.~\ref{fig:reflection}. Therefore, we leverage closed-source model APIs~\cite{hurst2024gpt,team2023gemini} and design a CoT image reflection annotation pipeline, enabling the models to step-by-step analyze image pairs. Specifically, we concatenate two images into an image grid and provide it as input together with the prompt. The model first identifies key differences between the two images, then makes a judgment regarding image preference, and finally produces a concise reflection instructing how to correct the identified flaws in the lower-quality image. The detailed CoT prompts are provided in Fig.~\ref{fig:cot-prompt-reflection} in the Appendix. 

Through this CoT-based annotation approach, we observe a significant improvement in the accuracy and reliability of generated reflections. Moreover, intermediate results from the reasoning process, such as the explicit image preferences, can also serve as valuable annotations for reward model training, as discussed in the next paragraph. We annotate approximately 270K CoT reflections using this annotation pipeline, and after careful filtering and quality control, we obtain a final dataset of 227K high-quality CoT reflections, named \dataset-CoT. Subsequently, we fine-tune the Qwen2.5-VL-7B model on this curated reflection dataset, enabling it to annotate the full dataset comprehensively\footnote{Appendix~\ref{app:dataset} shows a few samples from this dataset.}. Fig.~\ref{fig:reflection} illustrates a qualitative comparison among reflections generated by the original Qwen2.5-VL-7B, our fine-tuned reflector, and GPT-4o, clearly demonstrating that our fine-tuned model produces substantially improved and more accurate reflections compared to the original model. We also visualize the word cloud of reflections in Fig.~\ref{fig:data}, which shows the pattern of executable instructions.

\noindent\textbf{Verifier Post-processing.} To further ensure the quality of our dataset, particularly regarding the quality gap between paired images, we train an image reward model (verifier) to quantitatively evaluate image quality. Specifically, we leverage the intermediate image preference annotations from \dataset-CoT and selected pairs whose preferences align consistently with the image pair annotations in \dataset, serving as confident, high-quality preference data. Inspired by recent advancements in reward modeling for video generation~\cite{liu2025improving}, we adopt the Bradley–Terry (BT) pairwise comparison approach~\cite{bradley1952rank} to train our image reward model. The BT framework utilizes a pairwise log-likelihood loss to explicitly model the reward gap between preferred and non-preferred image pairs, defined as:
\begin{equation}
    \mathcal{L}_{\mathrm{BT}} = -\mathbb{E}_{(y, x_w, x_l)\sim D}\left[\log \sigma\left(r_{\eta}(x_w, y) - r_{\eta}(x_l, y)\right)\right],
\end{equation}
where \( y \) denotes the input prompt, \( (x_w, x_l) \) represents the preferred and non-preferred image pair respectively, \( r_{\eta} \) is the learnable reward model, and \( \sigma(\cdot) \) refers to the logistic sigmoid function. Leveraging this verifier, we conduct a rigorous post-processing step on our dataset, applying multiple criteria to filter out lower-quality data samples. Furthermore, the trained verifier can also be utilized for inference-time scaling, which we elaborate in detail in Section~\ref{exp:main}.

\subsection{Reflection Tuning}
\label{sec:training}

After having constructed a suitable dataset, we turn our attention to efficiently training a corrector model $C_{\phi}$ that improves the quality of generated images. Analogous to image editing tasks, we treat self-refinement as a conditional generation problem and employ an efficient fine-tuning strategy tailored for pretrained diffusion transformers (DiTs), without introducing any additional modules.

\noindent\textbf{Efficient Fine-tuning for DiT.} The MMDiT architecture~\cite{esser2024scaling} has become a de-facto for the recent T2I generation models. In these models, image tokens and textual embeddings are concatenated into a unified sequence, allowing joint multimodal attention within each transformer block. Inspired by recent advances in conditional generation~\cite{tan2024ominicontrol,xiao2024omnigen,li2025visualcloze}, we similarly concatenate textual inputs $y$, the flawed image $x_l$, and the refined image $x_w$ into a single fused sequence, enabling multimodal attention:

\begin{equation}
   \text{MMAttention}(z) = \text{softmax}\left(\frac{QK^\top}{\sqrt{d}}\right)V,
   \label{eq:attention}
\end{equation}
where the $Q$, $K$, and $V$ are linearly projected from the concatenated token sequence $z = [y; x_l; x_w]$, with $y$ being a concatenation of the original input prompt and the reflection prompt. This unified attention mechanism naturally facilitates bidirectional information exchange across multiple modalities without requiring specialized modules such as ControlNet~\cite{zhang2023adding} or IP-Adapter~\cite{ye2023ip}.

Under this framework, the flawed image $x_l$ serves as conditioning information and thus does not require applying the noise schedule and can be further downsampled to lower resolutions (from 1024 to 512) to boost computational efficiency during both training and inference. Consequently, we can directly apply standard diffusion (or flow-matching) loss~\cite{liu2022flow,lipman2022flow} on the refined target image $x_w$:

\begin{equation}
    \mathcal{L} = \mathbb{E}_{t, \epsilon, x_t}\left[\| C_{\phi}(x_t, t, y, x_l) - (x_w - \epsilon) \|_{2}^{2}\right] dt,
    \label{eq:loss}
\end{equation}
where the noise $\epsilon \sim \mathcal{N}(0, I)$ is sampled from a standard Gaussian distribution and $x_t$ is sampled from $p(x_t|\epsilon,x_w)$.

\noindent\textbf{Training Strategy.} We observe that directly training on our proposed dataset could cause distributional drift from the pretrained base model, subsequently degrading the quality of generated images. To mitigate this issue, we design a specialized training strategy to enhance model robustness and maintain alignment with the pretrained distribution. First, we randomly drop the original prompt, reflection prompt, and flawed input image with certain probabilities during training. The reflection prompt and flawed image are dropped simultaneously with a relatively high probability. In this way, the training objective effectively reverts to standard T2I generation, ensuring the model does not deviate excessively from its pretrained distribution.

Additionally, we adopt a ``task warm-up'' approach to dynamically adjust the sampling probabilities of multiple data domains throughout the training process. Initially, editing data is sampled with higher probability, facilitating rapid acquisition of accurate instruction-following capabilities. As training progresses, we gradually increase the sampling probabilities of the other three data domains. This gradual adjustment effectively improves the model's overall performance and compensates for potential visual quality degradation inherent in editing data by leveraging high-quality synthetic datasets.

\subsection{Test-Time Scaling via Iterative Refinement}
\label{sec:infer}

\begin{figure}[t]
    \centering
    \includegraphics[width=1.0\linewidth]{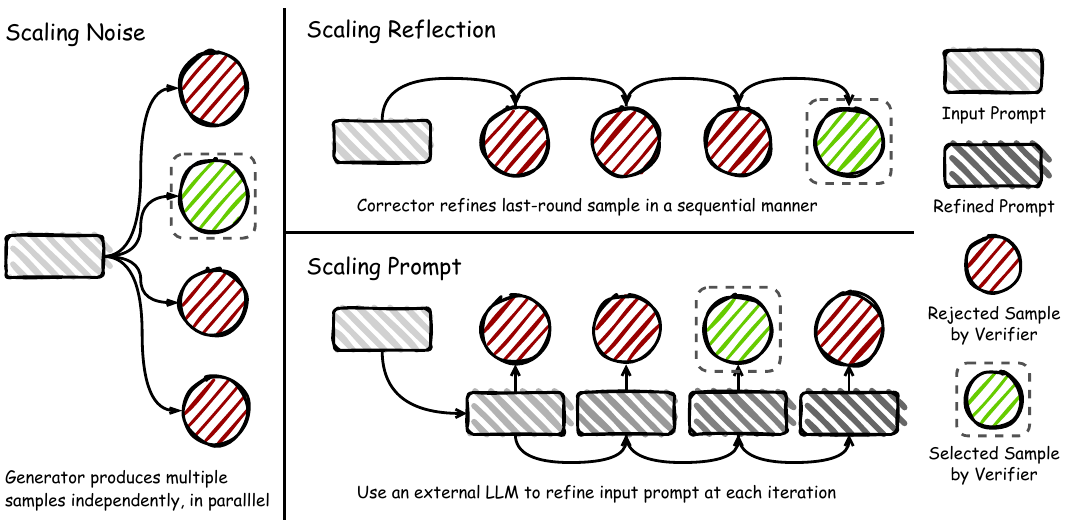}
    \caption{Illustrations of three different inference-time scaling dimensions for text-to-image diffusion models.}
    \label{fig:scaling}
\end{figure}

Leveraging the trained corrector model, we aim to maximize the generative capability of the diffusion model at inference time. In this section, we propose revisiting test-time scaling for T2I diffusion models along three distinct yet complementary dimensions: noise-level scaling, reflection-level scaling, and prompt-level scaling, as illustrated in Fig.~\ref{fig:scaling}. These three dimensions seamlessly integrate within our proposed \model framework. 

\input{Tabletex/mainresult}

\noindent\textbf{Noise-Level Scaling.} We first introduce noise-level scaling, inspired by recent works focused on noise-space optimization~\cite{zhou2024golden,ahn2024noise,qi2024not}. These approaches aim to identify superior initial noise or intermediate noisy images through various search strategies. Within our framework, we define the number of different initial noise samples explored per generation round as the search width $N$. By increasing $N$, one can fully explore the diversity embedded in the diffusion model's learned distribution. However, since the generation processes for these $N$ initial noise are independent and heavily reliant on feedback from a task-dependent verifier, simply scaling up $N$ can result in diminishing returns in terms of computational efficiency.

\noindent\textbf{Reflection-Level Scaling.} 
Reflection-level scaling builds upon our trained corrector model by iteratively refining previously generated images to progressively improve their quality. We define the number of iterative refinement rounds as the reflection depth $M$. Specifically, each refinement iteration follows the process defined in Section~\ref{sec:setup}, where the images generated in the previous iteration are refined iteratively with reflection. If the corrector model is effective, scaling the reflection depth $M$ can significantly enhance the model's overall performance. 

\noindent\textbf{Prompt-Level Scaling.} 
Finally, considering that T2I diffusion models rely not only on noisy image inputs but also on user-provided textual prompts, we design the prompt evolving process in our test-time scaling framework. We find that prompt expansion can substantially improve generation quality, especially for concise prompts. At each iterative round, we leverage an MLLM to refine the textual prompt based on the original user prompt, the previously generated images, and their evaluation scores. This prompt refinement procedure, performed without explicit gradient information, can produce precise and effective prompts for subsequent image generation iterations.

\noindent\textbf{\model.} Integrating all the aforementioned components, we propose \model framework, referring to Appendix~\ref{app:algo} for detailed algorithm. Specifically, we first employ a generator, implemented via offloading of LoRA weights, to produce an initial set of $N$ candidate images. Subsequently, at iteration $i$, we utilize an MLLM verifier to comprehensively evaluate and rank the $N$ images generated in the previous iteration across multiple dimensions. Based on these evaluation scores and previously generated images, the MLLM generates textual reflections aimed at correcting identified errors and then refines the user prompts. These reflections and improved prompts jointly serve as inputs to our trained corrector model with LoRA, producing a refined set of $N$ images for the next iteration. Finally, we use the verifier to select the best image for each refinement chain and then select the best image across all chains.

Our \model allows flexible adjustment of both the search width $N$ (number of parallel chains) and reflection depth $M$ (number of iterative refinement rounds). This flexibility empowers users to effectively balance performance and computational efficiency according to the requirements and constraints of various downstream tasks.

%% file: Tabletex/mainresult.tex
\begin{table*}[!t]
\centering
\vspace{-3mm}
\begin{tabular}{@{}c c | c | c | c c c c c c@{}}
\toprule
 & \textbf{Methods}  & \textbf{Samples} & \textbf{Overall} & \textbf{Single} & \textbf{Two} & \textbf{Counting} & \textbf{Colors} & \textbf{Position} & \textbf{Attribution} \\
\cmidrule{1-10}
\multicolumn{10}{c}{\cellcolor{grayred} \textbf{\textit{Text-to-Image Models}}} \\
\cmidrule{1-10}
& SDXL~\cite{podell2023sdxl} & 1 & 0.55  &  0.98 & 0.74 &  0.39 &  0.85 &  0.15 &  0.23 \\
& DALLE 3~\cite{betker2023improving} & 1 & 0.67  &  0.96 & 0.87 &  0.47 &  0.83 &  0.43 &  0.45 \\
& SANA-1.5 4.8B~\cite{xie2025sana} & 1 & 0.72 &  0.99 & 0.85 &  0.77 & 0.87  & 0.34  &  0.54  \\
& Lumina-Image 2.0~\cite{lumina2} & 1 & 0.73 &  0.99 & 0.87 &  0.67 & 0.88  & 0.34  &  0.62  \\
& SD3~\cite{esser2024scaling} & 1 & 0.74  &  0.99 & 0.94 &  0.72 &  0.89 &  0.33 &  0.60 \\
& Playground v3~\cite{liu2024playgroundv3improvingtexttoimage} & 1 & 0.76  &  0.99  & 0.95 & 0.72  & 0.82  &  0.50 & 0.54 \\
& Janus-Pro-7B~\cite{chen2025januspro} & 1 & 0.80  &  0.99  & 0.89 & 0.59  & 0.90  &  0.79 & 0.66 \\
\cmidrule{1-10}
\multicolumn{10}{c}{\cellcolor{uclagold} \textbf{\textit{Inference Time Scaling}}} \\
\cmidrule{1-10}

& SANA-1.0-1.6B~\cite{xie2024sana} & 1 & 0.66 &  0.99 & 0.77 &  0.62 & 0.88  & 0.21  &  0.47  \\
& + Noise Scaling\textsuperscript{\dag}~\cite{ma2025inference} & 20 & 0.80 &  1.00 & 0.93 &  0.79 & 0.91  & 0.55  &  0.62  \\ 
& + Reflect-DiT~\cite{li2025reflect} & 20 & 0.81 &  0.98 & 0.96 &  0.80 & 0.88  & 0.66  &  0.60  \\ 
\cmidrule{1-10}
& FLUX.1-dev~\cite{flux2024} & 1 & 0.67  &  0.99 & 0.81 &  0.75 &  0.80 &  0.21 &  0.48 \\
& + Noise Scaling\textsuperscript{\dag}~\cite{ma2025inference} & 32 & 0.85  & 1.00  & 0.96 & \textbf{0.91}  & 0.91  & 0.52  & \textbf{0.78} \\
& + Noise \& Prompt Scaling & 32 & 0.87 &  0.99 & 0.94 &  0.85 & 0.91  & 0.80  &  0.71  \\

& + \model & 32 & \textbf{0.91}  & \textbf{1.00} & \textbf{0.98} & 0.90 & \textbf{0.96}  & \textbf{0.93}  & 0.72 \\
\bottomrule
\end{tabular}
\caption{Quantitative comparisons of our \model framework against standard text-to-image models and different inference-time scaling approaches evaluated on the GenEval benchmark. The notation \textsuperscript{\dag} denotes results obtained using only noise-level scaling, which is equivalent to the random search strategy introduced in~\citet{ma2025inference}.}
\label{tab:mainresult}
\end{table*}

%% file: sec/4_exp.tex
\section{Experiments}

\begin{figure*}[t]
    \centering
    \includegraphics[width=1.0\linewidth]{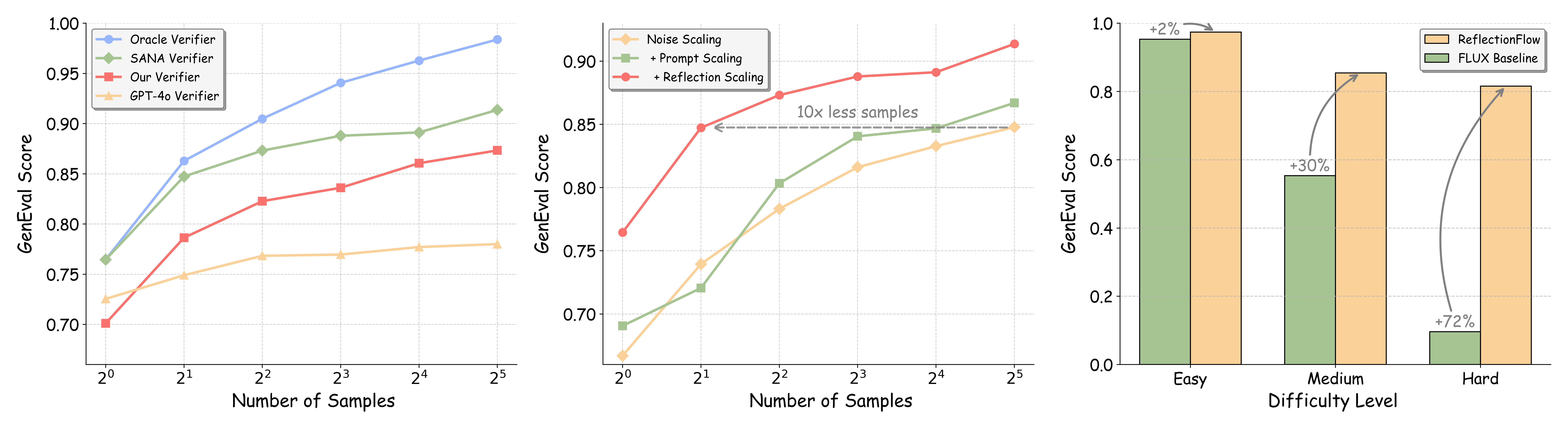}
    \caption{\textbf{Left:} The choice of verifier significantly impacts the effectiveness of inference-time scaling methods. \textbf{Middle:} By efficiently scaling the inference-time budget, \model achieves substantial performance improvements, requiring 10 times fewer samples compared to naive noise-level scaling. \textbf{Right:} \model demonstrates notably greater performance gains on challenging samples.}
    
    \label{fig:nfe}
\end{figure*}

\subsection{Setup}
\noindent\textbf{Training Details.}
We select FLUX.1-dev\footnote{\url{https://huggingface.co/black-forest-labs/FLUX.1-dev}}~\cite{flux2024}, a 12B-parameter flow-based diffusion transformer for text-to-image generation, as our base generator. To train the corrector model, we utilize our \dataset dataset, resizing and cropping all target images to $1024 \times 1024$ and all conditional images to $512 \times 512$. To enable efficient fine-tuning, we employ LoRA~\cite{hu2022lora} with a rank of 256. The model is trained using a batch size of 64 and optimized using the Prodigy optimizer~\cite{mishchenko2024prodigy}, with safeguard warmup and bias correction enabled, following practices outlined in OminiControl~\cite{tan2024ominicontrol}. The entire fine-tuning procedure is conducted over 6,000 optimization steps on 8 NVIDIA A100 GPUs.

For our reflection generator, we utilize Qwen2.5-VL-7B~\cite{Qwen2.5-VL} as the backbone model and apply full fine-tuning with a learning rate of $1 \times 10^{-6}$ for a total of 45,000 steps. Considering that the image reward modeling task is comparatively easier, we select Qwen2.5-VL-3B~\cite{Qwen2.5-VL} as the backbone of our verifier. Following the configuration of VideoAlign~\cite{liu2025improving}, we optimize the verifier using the Bradley-Terry (BT) loss, fine-tuning LoRA~\cite{hu2022lora} parameters of the language model and fully updating parameters of the vision encoder. Training is conducted for 10,080 steps with a learning rate of $2 \times 10^{-6}$.

\noindent\textbf{Evaluation.} We evaluate \model framework on GenEval benchmark~\cite{ghosh2024geneval}, following the official evaluation protocol. The GenEval dataset comprises 553 prompts, with four images generated per prompt. All images are generated at a resolution of $1024 \times 1024$, guidance scale of 3.5, and 30 sampling steps. We use the SANA verifier\footnote{\url{https://huggingface.co/Efficient-Large-Model/NVILA-Lite-2B-Verifier}} introduced in SANA-1.5~\cite{xie2025sana} to assess generated images in our main experiments, and use our fine-tuned Qwen2.5-VL-7B to produce reflection. We also test additional verifiers, including our verifier and GPT-4o, which is included in our verifier comparison experiments. The detailed GPT-4o prompt is provided in the Appendix.

\subsection{Main Results}
\label{exp:main}

We first validate the effectiveness of our proposed \model framework on the GenEval benchmark. Specifically, we progressively introduce three complementary scaling dimensions defined in our method: \emph{noise-level scaling}, \emph{prompt-level scaling}, and \emph{reflection-level scaling}. We compare \model with the baseline FLUX.1-dev and several state-of-the-art text-to-image models. Our main experiments are done under the setting of 32 samples.

\noindent\textbf{\model Boosts Generation Quality.} \cref{tab:mainresult} summarizes the quantitative results. Starting from the baseline FLUX.1-dev model (score of 0.67), we observe that introducing noise-level scaling significantly improves the score to 0.85. Further incorporating prompt-level scaling provides a slightly improvement, reaching a score of 0.87. Finally, when we integrate three levels of scaling together, \model can achieve a substantial leap in performance, reaching an overall score of 0.91.
These results demonstrate that naive noise-level scaling alone is inefficient without explicit guidance, while dynamically refining prompts along generation facilitates a clearer understanding of complex semantic attributes, such as spatial arrangements. The substantial performance gain achieved by incorporating reflection-level scaling further confirms that textual reflections from verifiers provide valuable feedback, enabling the corrector model to iteratively and reliably refine its outputs.

Moreover, we compare our method with a concurrent reflection-based work, Reflect-DiT~\cite{li2025reflect}. From \cref{fig:nfe}, either SANA verifier or our verifier's 16 samples results are better than Reflect-DiT's performance of 0.81, which is achieved with 20 samples per prompt. This comparison highlights the efficiency and effectiveness of our \model framework, which can be attributed to the better dataset, reflection model training, as well as the overall inference-time scaling framework.

\subsection{Ablation Studies}

\begin{figure*}[t]
    \centering
    \includegraphics[width=\textwidth]{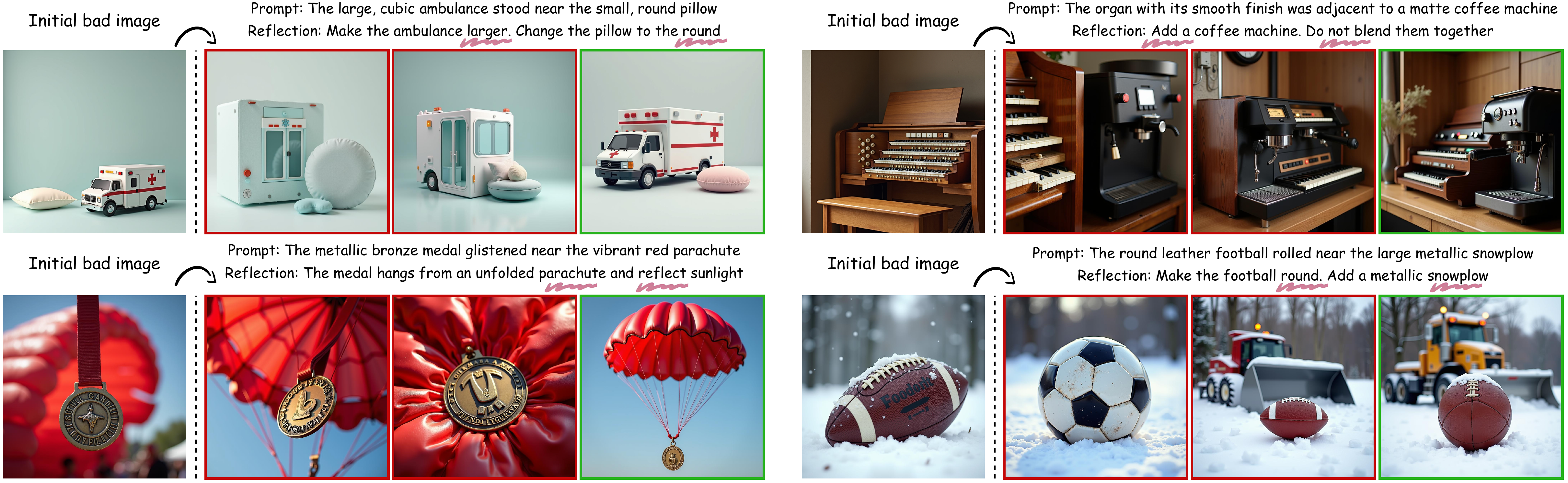}
    \caption{Visualization of complex reasoning. Starting from initially incorrect generations (the first image), \model iteratively reflects on and corrects errors, progressively producing images that accurately align with the provided prompts and reflection instructions .}
    \label{fig:complexreasoning}
\end{figure*}

\noindent\textbf{Exploring Different Verifiers.} 
To explore the potential of \model under various verifiers, we performed experiments using three distinct verifier settings: GPT-4o verifier, our fine-tuned verifier, and SANA verifier, which is specifically designed for the GenEval benchmark. Additionally, we utilized the oracle results from the GenEval evaluation pipeline to estimate the upper bound of current performance. The results are shown in \cref{fig:nfe}.

When using GPT-4o as the verifier, the performance curve quickly approaches its limit with only 32 samples per prompt. In contrast, when using our verifier trained with BT loss, we observe a steady increase in performance as the number of samples per prompt increases. Notably, this configuration does not yet reach the limit of inference-time scaling, indicating further potential for improvement.
Furthermore, by adopting the SANA verifier, which was designed for the GenEval dataset, we achieve even higher performance. Specifically, with 32 samples per prompt, our \model achieves a GenEval score of 0.91. For comparison, using the oracle results from the GenEval benchmark as an upper bound, we find that the model can achieve a score of up to 0.98, demonstrating the significant headroom for performance improvements with better verifiers.

These results show the strong inherent abilities of our \model, as its performance consistently improves with increasingly better verifiers with little sign of saturation. While the choice of verifier plays a critical role in unlocking this potential, the steady performance gains observed across different verifier settings highlight the robustness and scalability of our model. 

\noindent\textbf{Scaling Inference-time Budgets.}
We then investigate how the allocated inference-time budget influences the performance of our \model framework. Specifically, we fix the search width to 2 and progressively increase the reflection depth as we scale up the total budget. This allows us to explicitly examine the capability of \model to iteratively reflect and correct its previous mistakes.

Results shown in~\cref{fig:nfe} demonstrate that \model rapidly improves performance as the budget increases from 1 to 4, after which the improvement slows down, ultimately reaching a final GenEval score of 0.91 with 32 samples for each prompt. The baseline approaches, including ``Noise Scaling" and ``Noise \& Prompt Scaling" consistently underperform relative to \model across all budgets, indicating the limited effectiveness of simpler methods. We capped our evaluation at 32 samples for each prompt; however, the upward trajectory of \model's performance suggests that it has not yet reached its full potential, leaving room for further improvement with larger budgets. 

\input{Tabletex/brachdepth}

\noindent\textbf{Exploring Iterative Refinement Strategies.} 
We systematically investigate different exploration strategies within the \model framework. Specifically, we conduct ablation experiments by varying two key hyperparameters: search width $N$, which denotes the number of candidate images generated at each iteration, and reflection depth $M$, representing the number of iterative refinement rounds. Given a fixed computational budget of $N \times M = 16$, we explore three different refinement strategies: (1) \textit{Sequential}, generating one candidate per refinement step; (2) \textit{Parallel}, generating multiple candidates simultaneously per iteration; and (3) \textit{Combine}, balancing both depth and breadth by moderately expanding candidate branches per iteration. We use GPT-4o as the verifier.

~\cref{tab:bd_analysis} clearly shows that the sequential strategy (N1M16) achieves the highest overall performance of 0.78, outperforming the parallel strategy (N16M1), which yields a lower score of 0.74. Comparing various combined strategies, such as N2M8, N4M4, and N8M2, we consistently observe that strategies with greater refinement depth tend to yield better performance than those with wider branching. These observations suggest that \model framework exhibits effective reflection and self-correction capabilities, yet the refinement process is relatively unstable, requiring multiple sequential iterations to progressively identify and correct errors. Increasing the refinement depth allows the model to continuously reason and rectify previous mistakes, ultimately converging to improved results. 

\subsection{Analysis and Discussion}
\noindent\textbf{Reflection Capability for Difficult Tasks.}
Previous studies in LLMs demonstrate that inference-time scaling, particularly via longer reasoning and reflection chains,  improves performance on more challenging tasks~\cite{shinn2023reflexion,madaan2023self}. Inspired by these observations, we explore whether \model framework exhibits similar behavior in diffusion models. Leveraging the difficulty estimated in ~\cref{sec:dataset}, we divide the prompts from GenEval benchmark into three difficulty levels according to their initial correctness: \emph{hard} prompts with correctness between 0 and 0.3, \emph{medium} prompts with correctness between 0.4 and 0.7, and \emph{easy} prompts with correctness between 0.8 and 1.0. \cref{fig:nfe} summarizes the performance of \model across these three difficulty levels.

\model achieves the largest improvement on hard prompts, substantially increasing correctness from 0.10 to 0.81. Medium prompts exhibit moderate improvement, increasing correctness from 0.55 to 0.85, whereas easy prompts show minimal change, slightly increasing from 0.95 to 0.97. These results clearly indicate that \model shares a similar property with reflection-based scaling strategies in LLMs, where deeper iterative reflection is particularly beneficial for challenging tasks. This demonstrates the potential for future work to dynamically allocate inference-time compute based on prompt difficulty, leveraging the flexibility of \model.

\noindent\textbf{Qualitative Examples.} 
To intuitively illustrate the effectiveness of \model, we present qualitative examples of generated images and corresponding reflection processes. As shown in~\cref{fig:complexreasoning}, initial generations by our baseline often fail to capture detailed nuances or complex relations described in prompts. Through our iterative reflection, the model progressively identifies and corrects these issues, resulting in improved final outputs. Moreover, our method naturally produces interpretable reflection chains, which are similar to chain-of-thought reasoning in LLMs, demonstrating how the model explicitly reasons about and resolves visual inconsistencies, stylistic mismatches, and compositional inaccuracies step by step. Please refer to the Appendix for more qualitative results.


%% file: Tabletex/brachdepth.tex

\begin{table}[t]
\centering
\small
\begin{tabular}{@{}c c | c c c@{}}
\toprule
\textbf{Width} & \textbf{Depth} & \textbf{Overall} & \textbf{Position} & \textbf{Attribution} \\
\cmidrule{1-5}
16 & 1 & 0.74 & 0.56 & 0.42 \\
8 & 2 & 0.76 & 0.58 & 0.51 \\
4 & 4 & 0.77 & 0.56 & 0.51 \\
2 & 8 & 0.78 & 0.57 & \textbf{0.55} \\
1 & 16 & \textbf{0.78} & \textbf{0.69} & 0.51 \\
\bottomrule
\end{tabular}
\caption{Ablation studies on width and depth sizes in inference.}
\label{tab:bd_analysis}
\end{table}

%% file: sec/5_conclusion.tex
\section{Conclusion}

In this work, we present an inference-time framework, \model, that equips text-to-image diffusion models with iterative self-refinement capabilities. Central to our approach is the construction of \dataset, a large-scale image reflection dataset consisting of one million triplets of flawed images, high-quality images, and textual reflections. By formulating self-refinement as a generalized image editing problem and employing efficient fine-tuning strategies for pretrained diffusion transformers, \model effectively enhances performance without deviating from pretrained distributions. Furthermore, our framework integrates inference-time scaling across noise, reflection, and prompt dimensions, providing flexibility to balance computational efficiency and generation quality. We believe our work offers a promising direction for advancing self-refining generative models, contributing to more reliable and adaptive visual generation systems.

%% file: sec/6_appendix.tex
\ifpaper
\else

\section{Dataset Preview}
\label{app:dataset}

In the Figures \ref{fig:reward-samples}, \ref{fig:rule-samples}, and \ref{fig:length-samples}, we provide samples from our GenRef dataset. We divide these figures with respect to the subsets we used during data curation, except for the ``edit" samples that were sourced from the OmniEdit dataset \cite{wei2024omniedit}. For each image, we provide the prompt and reflection pairs. The red and green borders indicate the starting and final images, respectively. For best viewing experience, we recommend zooming in.

\begin{figure}[htbp]
  \centering
  \begin{subfigure}[b]{0.32\textwidth}
    \centering
    \includegraphics[width=\textwidth]{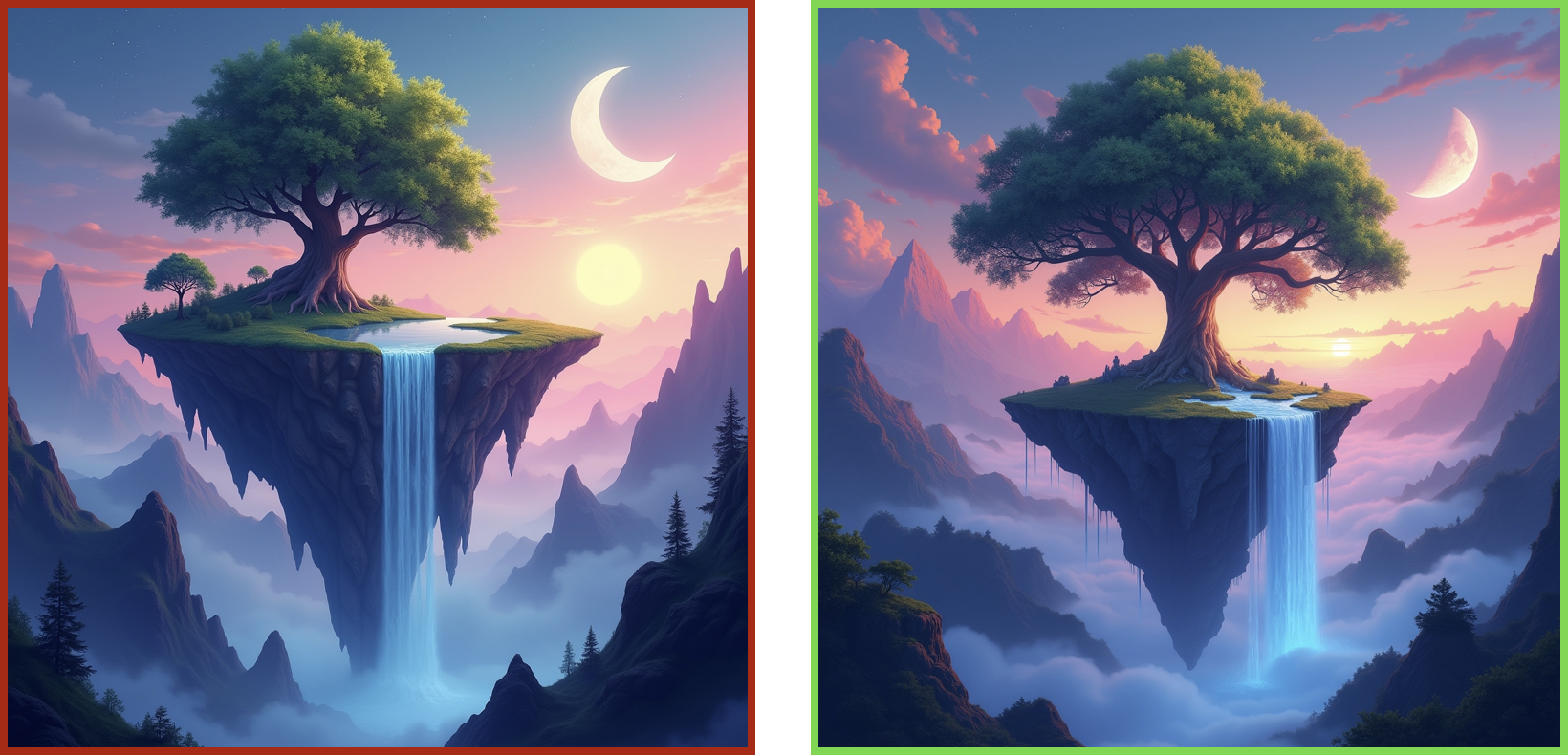}
    \caption{}
    \label{fig:reward-sub-a}
  \end{subfigure}\hfill
  \begin{subfigure}[b]{0.32\textwidth}
    \centering
    \includegraphics[width=\textwidth]{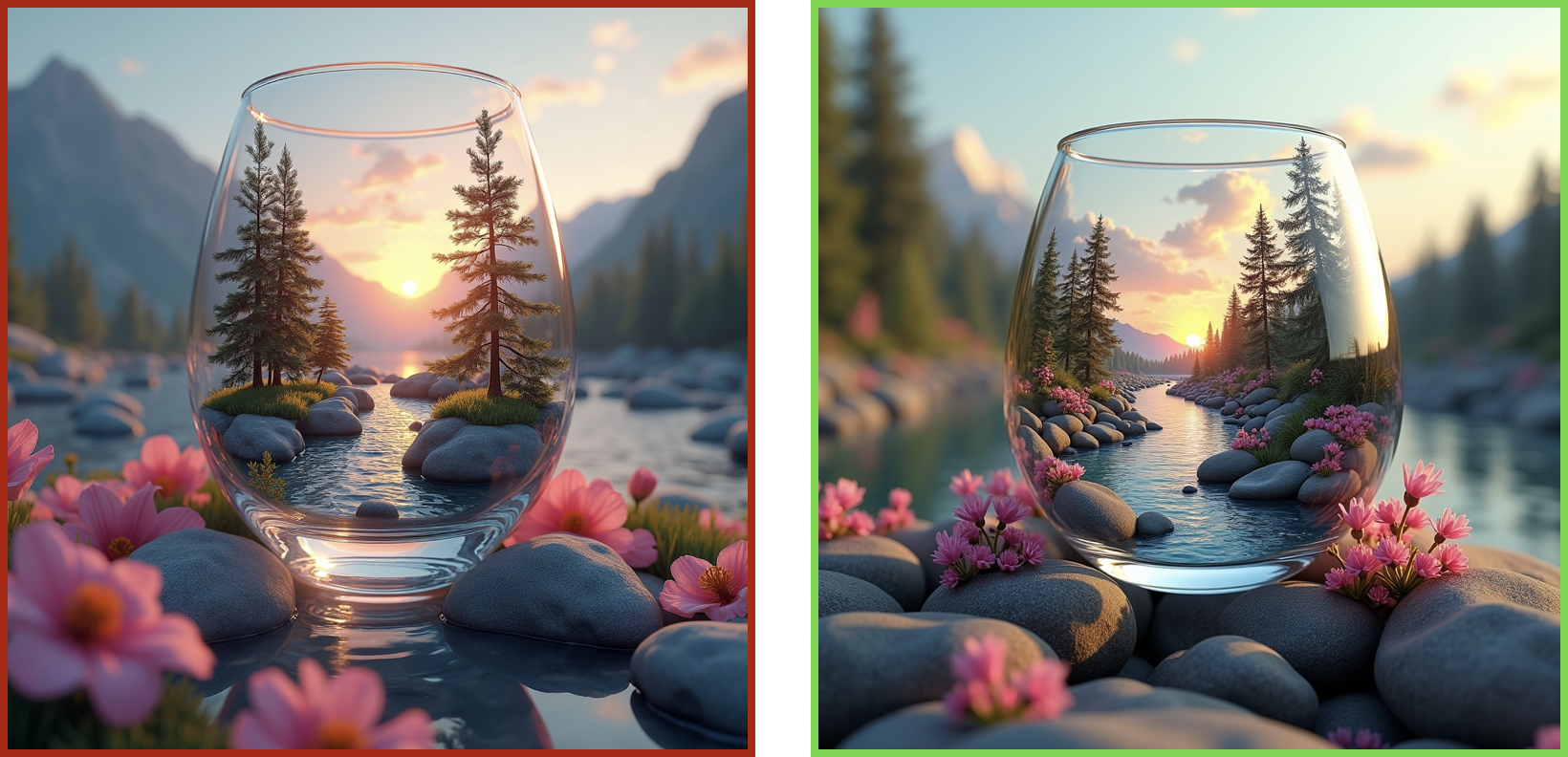}
    \caption{}
    \label{fig:reward-sub-b}
  \end{subfigure}\hfill
  \begin{subfigure}[b]{0.32\textwidth}
    \centering
    \includegraphics[width=\textwidth]{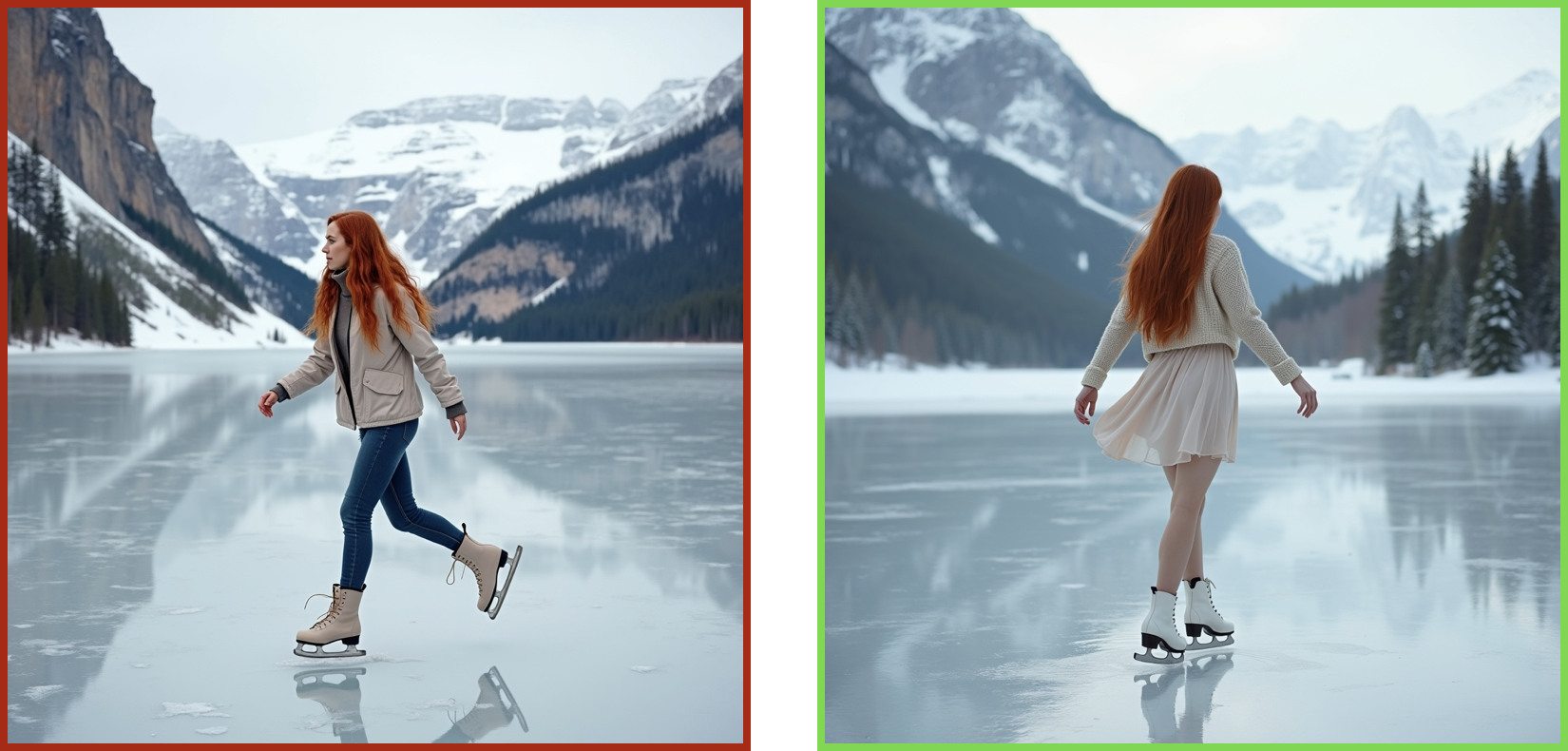}
    \caption{}
    \label{fig:reward-sub-c}
  \end{subfigure}

  \caption{Samples from the pool of reward-based data. (\textbf{a}) \textbf{Prompt}: \textit{A surreal digital illustration of a floating island with a large, lush tree and cascading waterfalls, set against a twilight sky with a crescent moon, surrounded by misty mountains and vibrant, ethereal colors}. \textbf{Reflection}: \textit{Remove the sun in the background. Add more mist around the mountains}. (\textbf{b}) \textbf{Prompt}: \textit{A surreal digital artwork depicting a clear glass resting on rocks, containing a miniature landscape with a river, pine trees, and a sunset, surrounded by pink flowers, creating a dreamlike, photorealistic scene with vibrant colors and intricate details}. \textbf{Reflection}: \textit{Add more rocks around the glass. Add more pink flowers around the glass}. (\textbf{c}) \textbf{Prompt}: \textit{A young woman with long red hair ice skates gracefully on a frozen lake, surrounded by snow-covered mountains and evergreen trees, creating a serene and ethereal winter scene}. \textbf{Reflection}: \textit{Change the woman's outfit to a white sweater and skirt. Make the woman skate more gracefully}.}
  \label{fig:reward-samples}
\end{figure}

\begin{figure}[htbp]
  \centering
  \begin{subfigure}[b]{0.32\textwidth}
    \centering
    \includegraphics[width=\textwidth]{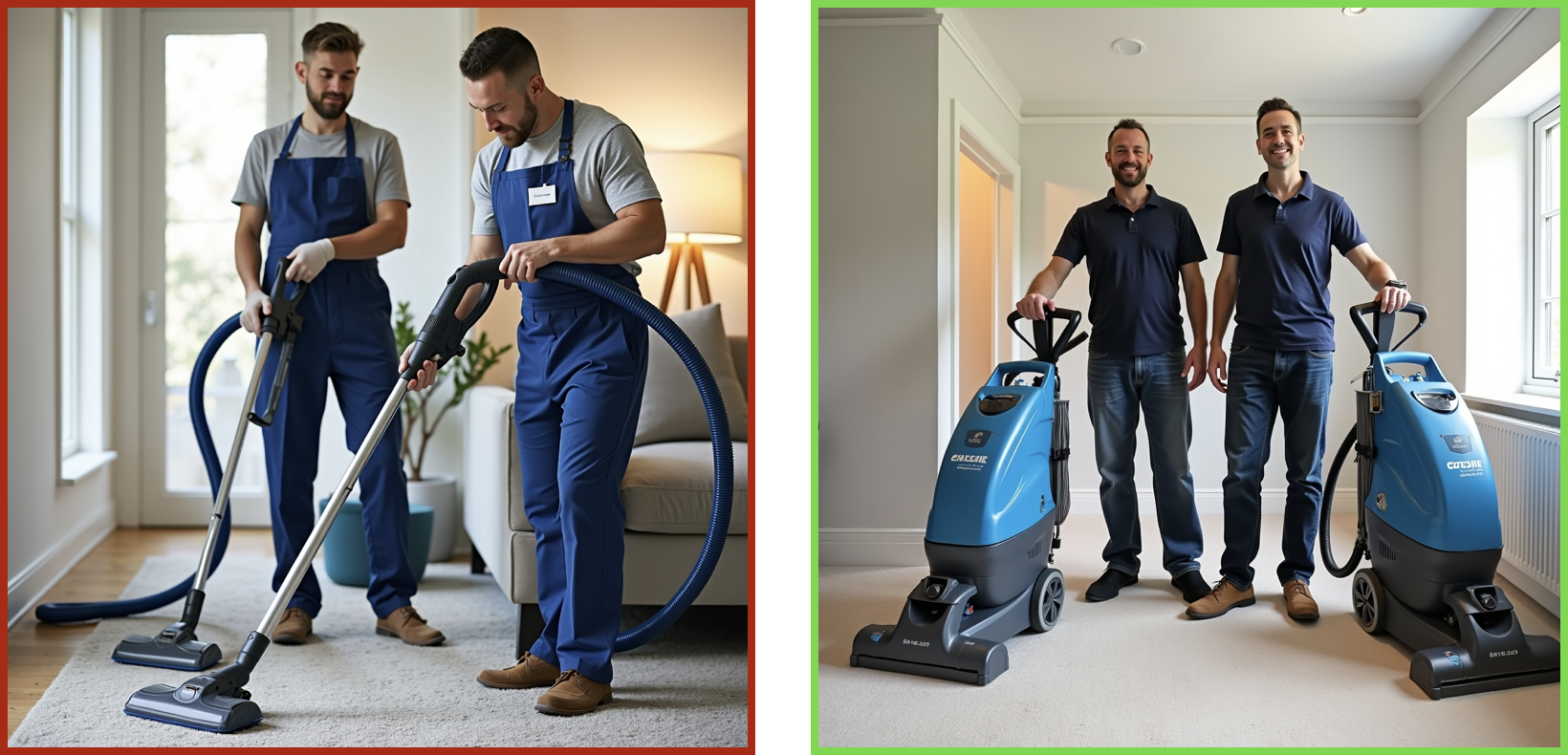}
    \caption{}
    \label{fig:rule-sub-a}
  \end{subfigure}\hfill
  \begin{subfigure}[b]{0.32\textwidth}
    \centering
    \includegraphics[width=\textwidth]{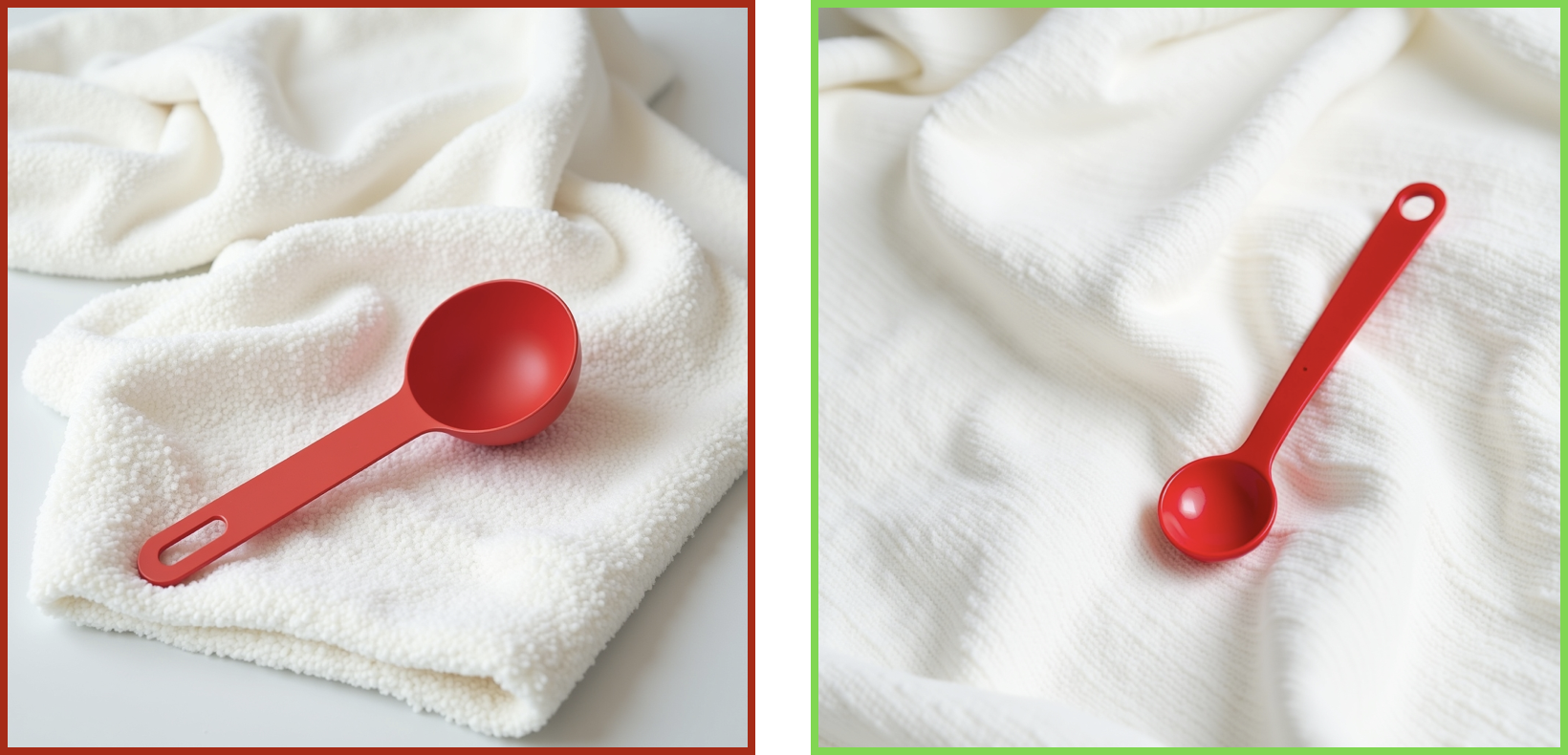}
    \caption{}
    \label{fig:rule-sub-b}
  \end{subfigure}\hfill
  \begin{subfigure}[b]{0.32\textwidth}
    \centering
    \includegraphics[width=\textwidth]{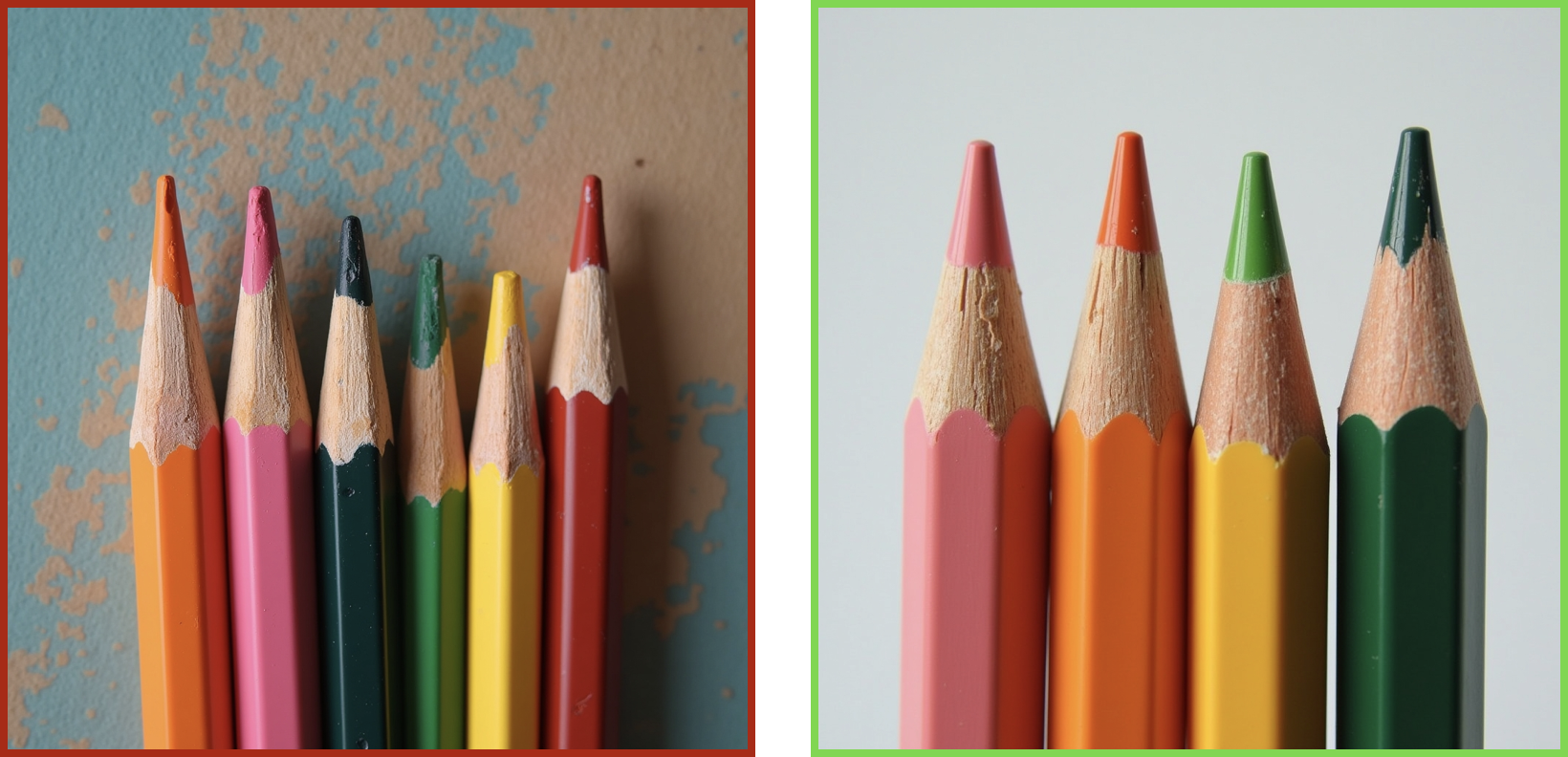}
    \caption{}
    \label{fig:rule-sub-c}
  \end{subfigure}

  \caption{Samples from the pool of rule-based data. (\textbf{a}) \textbf{Prompt}: \textit{a photo of two carpet cleaners}. \textbf{Reflection}: \textit{Replace the vacuum cleaners with professional-grade carpet cleaning machines and adjust the posture of the individuals to face forward while holding the machines}. (\textbf{b}) \textbf{Prompt}: \textit{a photo of a white blanket and a red measuring spoon}. \textbf{Reflection}: \textit{Change the texture of the blanket to a smooth, woven pattern instead of a fluffy one}. (\textbf{c}) \textbf{Prompt}: \textit{a photo of four colored pencils}. \textbf{Reflection}: \textit{Remove two pencils from the group to leave only four pencils visible}.}
  \label{fig:rule-samples}
\end{figure}

\begin{figure}[htbp]
  \centering
  \begin{subfigure}[b]{0.32\textwidth}
    \centering
    \includegraphics[width=\textwidth]{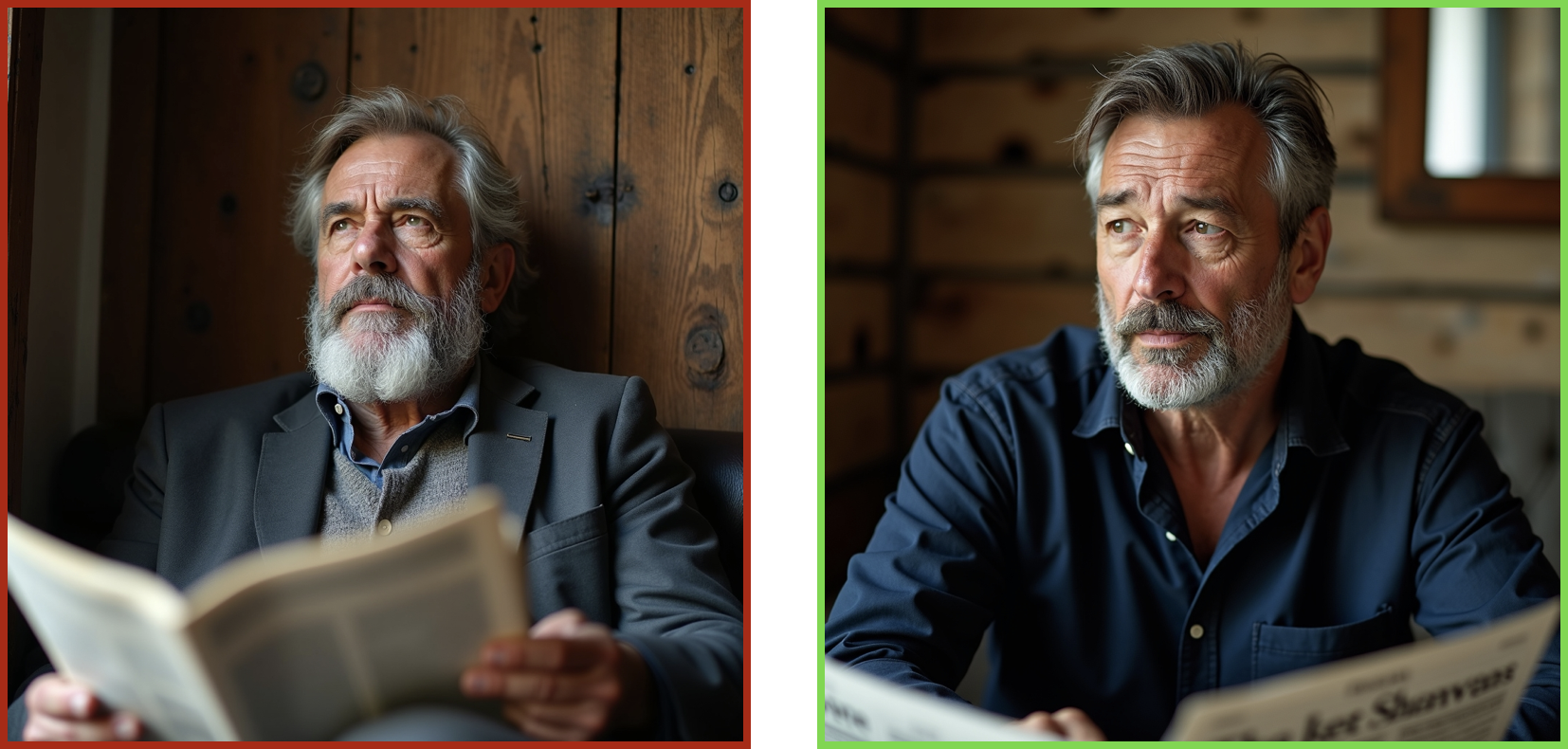}
    \caption{}
    \label{fig:length-sub-a}
  \end{subfigure}\hfill
  \begin{subfigure}[b]{0.32\textwidth}
    \centering
    \includegraphics[width=\textwidth]{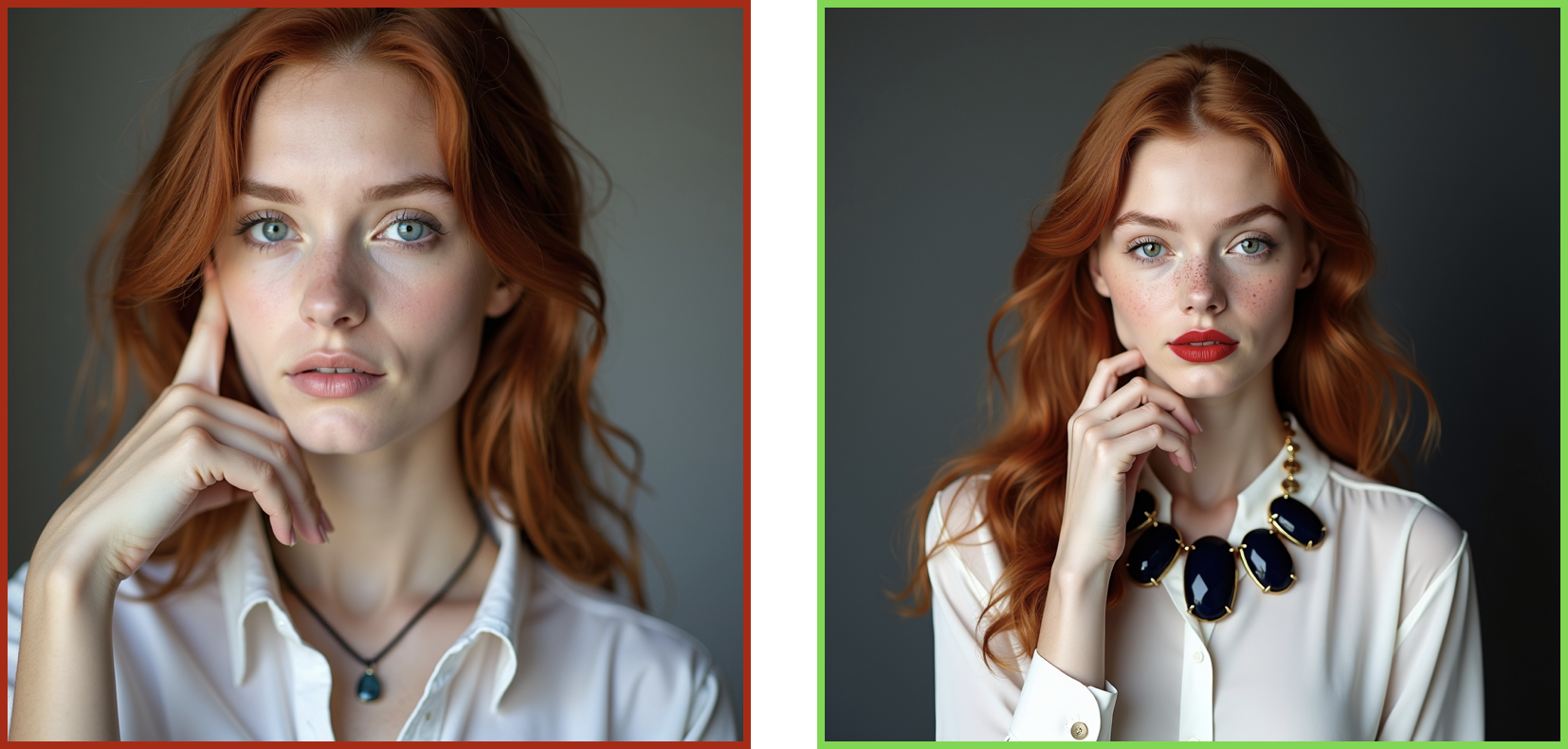}
    \caption{}
    \label{fig:length-sub-b}
  \end{subfigure}\hfill
  \begin{subfigure}[b]{0.32\textwidth}
    \centering
    \includegraphics[width=\textwidth]{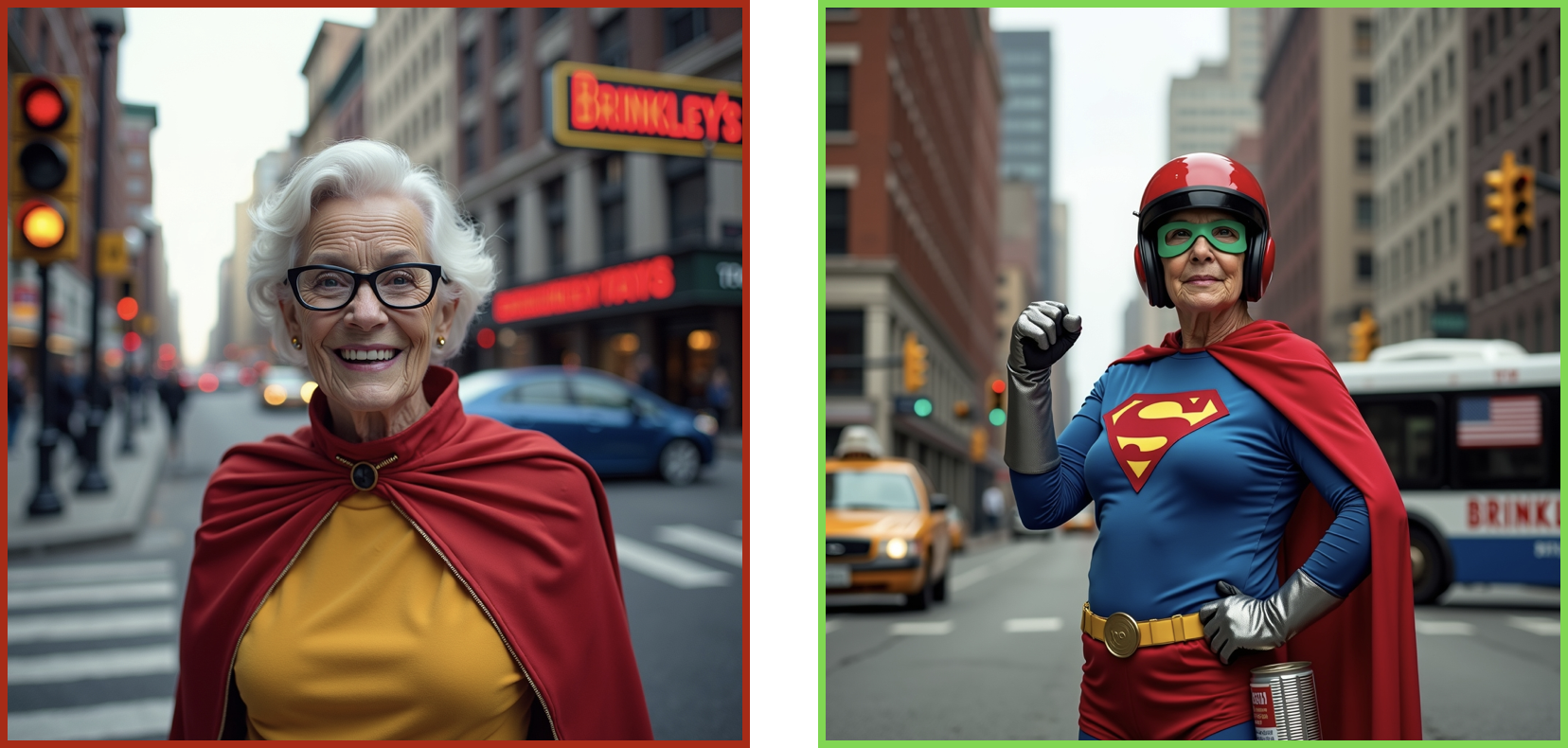}
    \caption{}
    \label{fig:length-sub-c}
  \end{subfigure}

  \caption{Samples from the pool of long-short prompt data. (\textbf{a}) \textbf{Prompt}: \textit{Portrait of a middle-aged man with a beard, seated indoors, looking slightly to the right. He wears a dark blue shirt and is positioned in the lower left of the frame. His right hand holds a newspaper, partially visible in the foreground. The background features rustic wooden walls with a warm, weathered texture, and a wooden mirror frame is partially visible on the right. The lighting is soft and diffused, casting gentle shadows on his face, creating a contemplative mood. The color palette is muted with earthy tones, emphasizing a cozy, intimate atmosphere. The composition is balanced, with a shallow depth of field that keeps the focus on the man's expression. Photorealistic, cinematic, warm, introspective, visually balanced}. \textbf{Reflection}: \textit{Change the suit jacket into a dark blue shirt. Remove the gray sweater vest}. (\textbf{b}) \textbf{Prompt}: \textit{Studio portrait of a young woman with fair skin and long, wavy red hair, centered against a dark grey background. She gazes directly at the camera with a neutral expression, her lips painted a vibrant red. Her right hand is raised to her chin, with fingers gently touching her cheek. She wears a crisp white blouse with a statement necklace featuring large, dark blue gemstones. The lighting is soft and even, highlighting her freckles and the texture of her hair. High contrast, sharp focus, professional studio photography, neutral color palette, elegant and poised, classic portrait composition}. \textbf{Reflection}: \textit{Paint the lips a vibrant red. Replace the necklace with one that features large, dark blue gemstones}. (\textbf{c}) \textbf{Prompt}: \textit{A striking portrait of an elderly woman dressed as a superhero in an urban setting. She stands confidently in the foreground, wearing a red helmet with a visor, green eye mask, and a red and blue superhero costume with a cape. Her right hand is raised, adorned with a silver glove, while her left hand rests on her hip, also gloved. A can of food is strapped to her left arm. The background features a city street with a yellow traffic light and a bus with an American flag on its side, parked on the right. A brick building with the sign ``BRINKLEY'S" is visible behind her. The scene is set in a bustling city environment with blurred buildings and a taxi in the distance. Photorealistic, high contrast, dramatic lighting, vibrant color palette, sharp focus on the subject, urban superhero theme, dynamic composition, slightly desaturated background, cinematic feel}. \textbf{Reflection}: \textit{Change the woman's outfit to a red and blue superhero costume with a cape. Add a red helmet with a visor, green eye mask, and a silver glove to the woman's hand}.}
  \label{fig:length-samples}
\end{figure}

\noindent Figure \ref{fig:cot-samples} shows samples from the dataset created for fine-tuning our Qwen model, as described in Section \ref{sec:dataset}. The green and red borders denote the correct and incorrect images (as deemed by closed-source APIs), respectively.

\begin{figure}[htbp]
  \centering
  \begin{subfigure}[b]{0.32\textwidth}
    \centering
    \includegraphics[width=\textwidth]{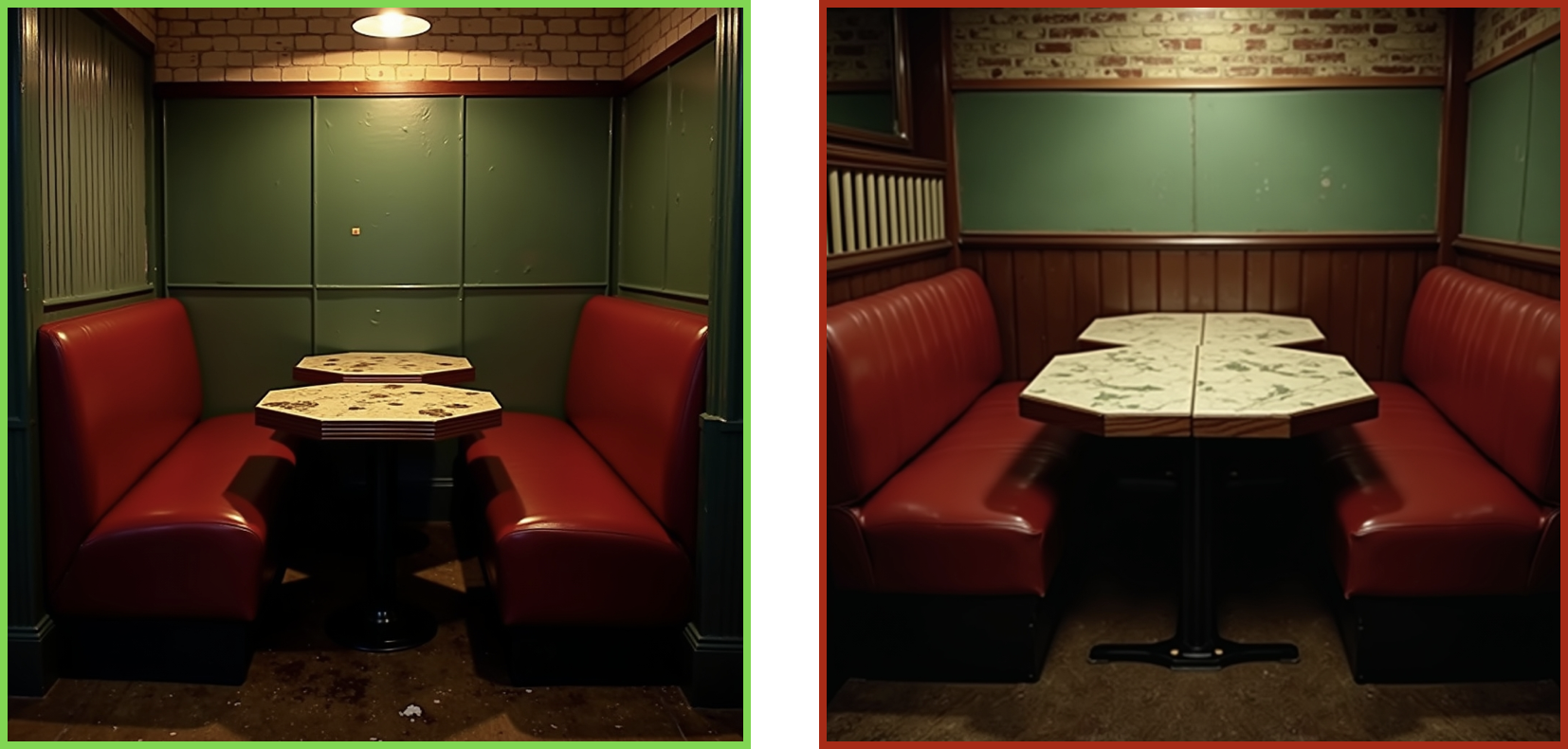}
    \caption{}
    \label{fig:cot-sub-a}
  \end{subfigure}\hfill
  \begin{subfigure}[b]{0.32\textwidth}
    \centering
    \includegraphics[width=\textwidth]{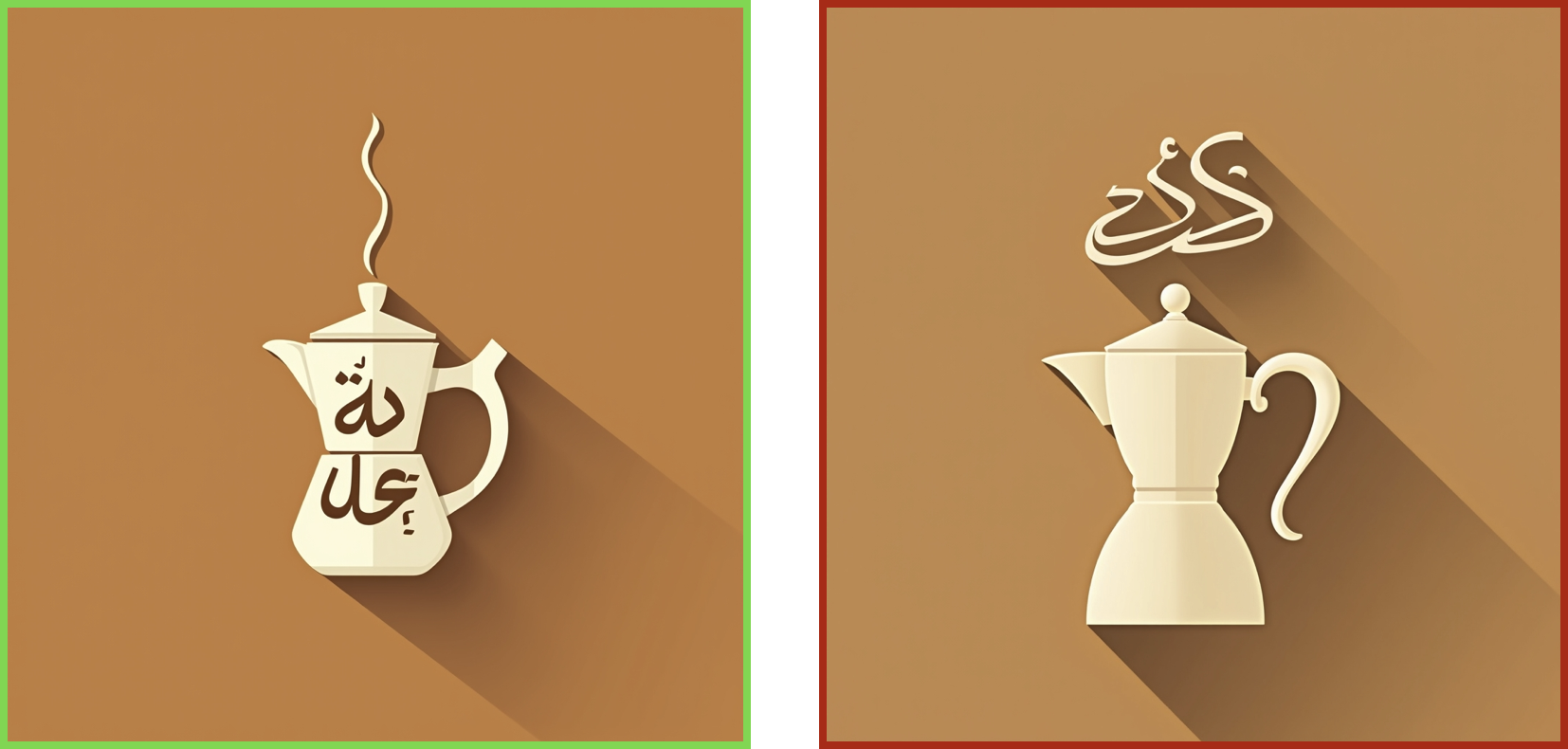}
    \caption{}
    \label{fig:cot-sub-b}
  \end{subfigure}\hfill
  \begin{subfigure}[b]{0.32\textwidth}
    \centering
    \includegraphics[width=\textwidth]{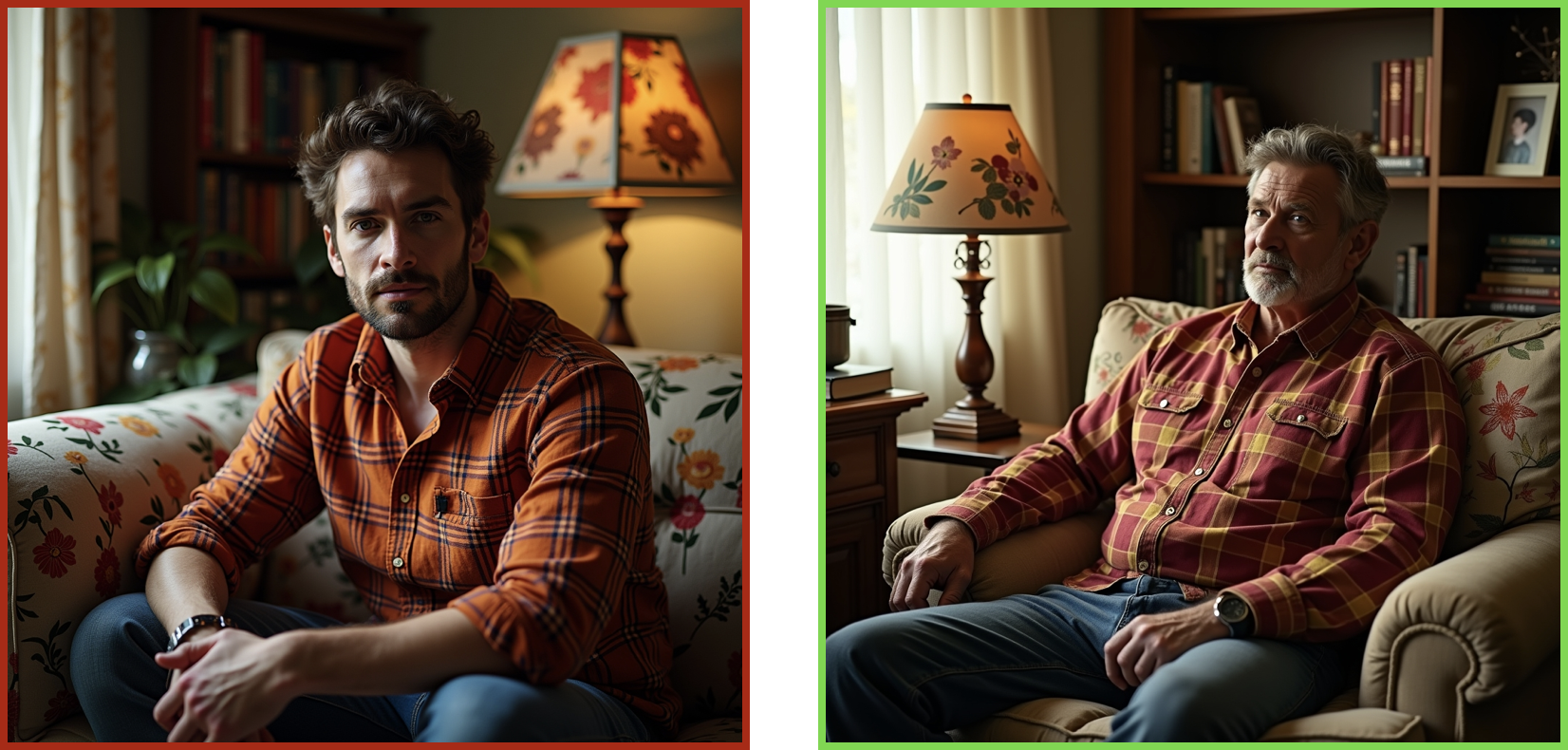}
    \caption{}
    \label{fig:cot-sub-c}
  \end{subfigure}

  \caption{Samples from the pool of CoT (chain-of-thought) data. (\textbf{a}) \textbf{Prompt}: \textit{Dimly lit photograph of an empty booth in a cozy, vintage-style diner. Red leather benches form an L-shape around a pair of adjoining hexagonal marble tables in the left center against the back wall. The tables, with dark veining and brown wooden edges, sit atop black metal bases. The wall shows a textured, weathered green paint over a brick surface, subtly shadowed. In the foreground, a waist-high partition with vertical ribbed glass edges frames the scene on the right. Flooring includes worn terrazzo tiles with visible dust. Warm, ambient lighting creates an inviting yet mysterious atmosphere. Photorealistic, shallow depth of field, warm tones, nostalgic, intimate mood, reminiscent of mid-20th-century design}. \textbf{Analysis}: \textit{The left image better matches the prompt. It features a dimly lit, cozy atmosphere with red leather benches forming an L-shape around two adjoining hexagonal marble tables. The wall has a textured green paint over a brick surface, and the terrazzo flooring appears worn and dusty. The right image deviates from the prompt as the benches form a U-shape instead of an L-shape, the wall includes wooden paneling not mentioned in the prompt, and the lighting is less dim and warm, reducing the nostalgic and intimate mood}. \textbf{Instruction}: \textit{Change the U-shaped bench arrangement to an L-shape. Remove the wooden paneling on the walls and replace it with textured green paint over a brick surface}. (\textbf{b}) \textbf{Prompt}: \textit{A minimalist, flat-design illustration of stylized Arabic calligraphy forming a coffee pot with steam, set against a warm brown background, featuring a cream color palette and subtle shadows for depth}. \textbf{Analysis}: \textit{The left image accurately portrays stylized Arabic calligraphy forming a coffee pot and steam, while the right image only contains Arabic calligraphy around a regular coffee pot. The prompt specifically calls for the Arabic calligraphy to form the pot and steam}. \textbf{Instruction}: \textit{Transform the Arabic calligraphy into the form of a coffee pot with steam. Remove the existing regular coffee pot}. (\textbf{c}) \textbf{Prompt}: \textit{A man in a red and yellow plaid shirt sits on a floral sofa in a warmly lit, cozy living room with natural light, a bookshelf, and a floral lamp, creating a relaxed and intimate atmosphere}. \textbf{Analysis}: \textit{The right image is better because it depicts a man in a red and yellow plaid shirt seated on a floral sofa within a warmly lit, cozy living room, effectively conveying a relaxed and intimate setting as described in the prompt. The left image features a man in a plaid shirt, but the colors are not quite right, and the setting is not as warm or cozy, and the lighting is not right}. \textbf{Instruction}: \textit{Change the shirt color to red and yellow plaid. Add natural light to give the setting a warm and cozy feel}.}
  \label{fig:cot-samples}
\end{figure}

\clearpage

\section{Qualitative Results}
In Figure \ref{fig:quantitative}, we list some qualitative results of our ReflectionFlow framework. This includes the
detailed process of our three scaling steps, giving a more nuanced understanding of our framework.

\begin{figure}[H]
    \centering
    \includegraphics[width=\textwidth]{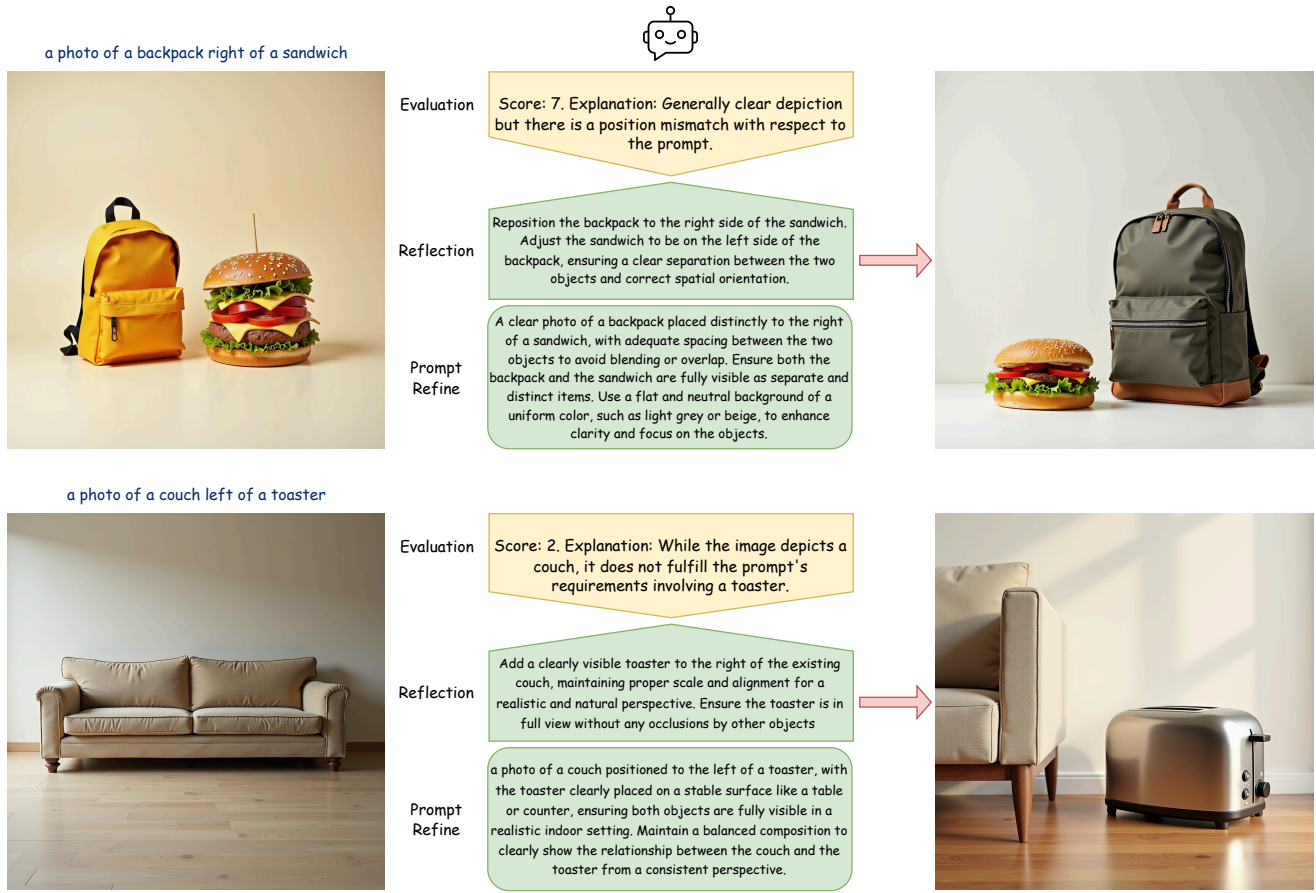}
    \caption{Qualitative results involving complex reasoning.}
    \label{fig:quantitative}
\end{figure}

\clearpage

\section{Algorithm Process}
\label{app:algo}
The proposed ReflectionFlow framework is as follows:

\begin{algorithm}[H]
\caption{The proposed ReflectionFlow framework}
\label{alg:model}
\begin{algorithmic}[1]
\Require prompt $y$, generator $G_\theta$, corrector $C_\phi$, verifier $V$, scaling width $N$, scaling depth $M$
\Ensure High-quality image that best realizes user intent
\State $X_0 \gets \emptyset$ \Comment{Initial image set}
\For{$j = 1$ to $N$}
    \State Sample $z^j \sim \mathcal{N}(0, I)$
    \State $x_0^j \gets G_\theta(y, z^j)$ \Comment{Generate initial image}
    \State $X_0 \gets X_0 \cup \{x_0^j\}$
\EndFor
\For{$i = 1$ to $M$}
    \State $s_i \gets V(X_{i-1}, y)$ \Comment{Score previous images}
    \State $X_i \gets \emptyset$ 
    \For{$j = 1$ to $N$}
        \State $r_i^j, y_i^j \gets V(x_{i-1}^j, y, s_i)$ \Comment{Generate reflection}
        \State $x_i^j \gets C_\phi(y_i^j, r_i^j, x_{i-1}^j)$ \Comment{Refine with corrector}
        \State $X_i \gets X_i \cup \{x_i^j\}$
    \EndFor
\EndFor
\State \Return $\arg\max_{x \in \bigcup_{i=0}^{M}\bigcup_{j=1}^{N}{x_{i}^j}} V(x, y)$
\end{algorithmic}
\end{algorithm}

\clearpage

\section{Prompts}
We provide all the prompts we used throughout this work. These prompts were inspired by Figure 16 from \cite{ma2025inference}.
\subsection{Verifier Prompts}
\subsubsection{Single Object}
\begin{figure}[H]
\centering
\fbox{%
\begin{minipage}{1.0\textwidth}
\ttfamily
You are a multimodal large-language model tasked with evaluating images generated by a text-to-image model. Your goal is to assess each generated image based on specific aspects and provide a detailed critique, along with a scoring system. The final output should be formatted as a JSON object containing individual scores for each aspect and an overall score. The keys in the JSON object should be: \texttt{object\_completeness}, \texttt{detectability}, \texttt{occlusion\_handling}, and \texttt{overall\_score}. Below is a comprehensive guide to follow in your evaluation process:

\textbf{1. Key Evaluation Aspects and Scoring Criteria:}

For each aspect, provide a score from 0 to 10, where 0 represents poor performance and 10 represents excellent performance. For each score, include a short explanation or justification (1-2 sentences) explaining why that score was given. The aspects to evaluate are as follows:

\textbf{a) Object Completeness:}

Assess the structural integrity of the object (no defects/deformations), detail clarity and legibility.
Score: 0 (severely fragmented) to 10 (perfectly intact).

\textbf{b) Detectability:}

Evaluate the distinction and visual saliency of objects and backgrounds using contrast analysis.
Score: 0 (camouflaged) to 10 (immediately noticeable).

\textbf{c) Occlusion Handling:}

Assess whether there is unreasonable occlusion (natural occlusion needs to keep the subject visible).
Score: 0 (key parts are blocked) to 10 (no blockage/natural and reasonable blockage).

\textbf{2. Overall Score}:

After scoring each aspect individually, provide an overall score, representing the model's general performance on this image. This should be a weighted average based on the importance of each aspect to the prompt or an average of all aspects.
\end{minipage}
}
\caption{Verifier prompt for images with single object.}
\label{fig:gpt4-prompt-single-object}
\end{figure}
\clearpage

\subsubsection{Two Objects}
\begin{figure}[H]
\centering
\fbox{%
\begin{minipage}{1.0\textwidth}
\ttfamily
You are a multimodal large-language model tasked with evaluating images generated by a text-to-image model. Your goal is to assess each generated image based on specific aspects and provide a detailed critique, along with a scoring system. The final output should be formatted as a JSON object containing individual scores for each aspect and an overall score. The keys in the JSON object should be: \texttt{separation\_clarity}, \texttt{individual\_completeness}, \texttt{relationship\_accuracy}, and \texttt{overall\_score}. Below is a comprehensive guide to follow in your evaluation process: Your evaluation should focus on these aspects:

\textbf{1. Key Evaluation Aspects and Scoring Criteria}:
For each aspect, provide a score from 0 to 10, where 0 represents poor performance and 10 represents excellent performance. For each score, include a short explanation or justification (1-2 sentences) explaining why that score was given. The aspects to evaluate are as follows: 

\textbf{a) Seperation Clarity}:
Assess the spatial separation and boundary clarity of two objects.
Score: 0 (fully overlapped) to 10 (completely separate and clearly defined boundaries)

\textbf{b) Indivisual Completeness}:
Evaluate each object's individual integrity and detail retention.
Score: 0 (both objects are incomplete) to 10 (both objects are complete).

\textbf{c) Relationship Accuracy}:
Assess the rationality of size proportions.
Score: 0 (wrong proportions) to 10 (perfectly in line with physical laws).

\textbf{2. Overall Score}: 
After scoring each aspect individually, provide an overall score, representing the model's general performance on this image. This should be a weighted average based on the importance of each aspect to the prompt or an average of all aspects.
\end{minipage}
}
\caption{Verifier prompt for images with two objects.}
\label{fig:gpt4-prompt-two-objects}
\end{figure}
\clearpage
\subsubsection{Counting}
\begin{figure}[H]
\centering
\fbox{%
\begin{minipage}{1.0\textwidth}
\ttfamily
You are a multimodal large-language model tasked with evaluating images generated by a text-to-image model. Your goal is to assess each generated image based on specific aspects and provide a detailed critique, along with a scoring system. The final output should be formatted as a JSON object containing individual scores for each aspect and an overall score. The keys in the JSON object should be: \texttt{count\_accuracy}, \texttt{object\_uniformity}, \texttt{spatial\_legibility}, and \texttt{overall\_score}. Below is a comprehensive guide to follow in your evaluation process: Your evaluation should focus on these aspects:

\textbf{1. Key Evaluation Aspects and Scoring Criteria}:
For each aspect, provide a score from 0 to 10, where 0 represents poor performance and 10 represents excellent performance. For each score, include a short explanation or justification (1-2 sentences) explaining why that score was given. The aspects to evaluate are as follows: 

\textbf{a) Count Accuracy}:
Assess the number of generated objects matches the exact prompt.
Score: 0 (number wrong) to 10 (number correct).

\textbf{b) Object Uniformity}:
Evaluate the consistency of shape/size/color among same kind of objects.
Score: 0 (same kind but total different shape/size/color) to 10 (same kind and same shape/size/color).

\textbf{c) Spatial Legibility}:
Evaluate the plausibility and visibility of object distribution (no excessive overlap).
Score: 0 (heavily overlapped) to 10 (perfect displayed and all easily seen).

\textbf{2. Overall Score}: 
After scoring each aspect individually, provide an overall score, representing the model's general performance on this image. This should be a weighted average based on the importance of each aspect to the prompt or an average of all aspects.
\end{minipage}
}
\caption{Verifier prompt for images for counting.}
\label{fig:gpt4-prompt-counting}
\end{figure}
\clearpage
\subsubsection{Colors}
\begin{figure}[H]
\centering
\fbox{%
\begin{minipage}{1.0\textwidth}
\ttfamily
You are a multimodal large-language model tasked with evaluating images generated by a text-to-image model. Your goal is to assess each generated image based on specific aspects and provide a detailed critique, along with a scoring system. The final output should be formatted as a JSON object containing individual scores for each aspect and an overall score. The keys in the JSON object should be: \texttt{color\_fidelity}, texttt{contrast\_effectiveness}, \texttt{multi\_object\_consistency}, and \texttt{overall\_score}. Below is a comprehensive guide to follow in your evaluation process: Your evaluation should focus on these aspects:

\textbf{1. Key Evaluation Aspects and Scoring Criteria}:
For each aspect, provide a score from 0 to 10, where 0 represents poor performance and 10 represents excellent performance. For each score, include a short explanation or justification (1-2 sentences) explaining why that score was given. The aspects to evaluate are as follows: 

\textbf{a) Color Fidelity}:
Assess the exact match between the object color and the input prompt.
Score: 0 (color wrong) to 10 (color correct).

\textbf{b) Contrast Effectiveness}:
Evaluate the difference between foreground and background colors.
Score: 0 (similar colors, difficult to distinguish) to 10 (high contrast).

\textbf{c) Multi-Object Consistency}:
Assess color consistency across multiple same kind of objects.
Score: 0 (same kind of objects with total different colors) to 10 (same kind with same color).

\textbf{2. Overall Score}: 
After scoring each aspect individually, provide an overall score, representing the model's general performance on this image. This should be a weighted average based on the importance of each aspect to the prompt or an average of all aspects.
\end{minipage}
}
\caption{Verifier prompt for images for colors.}
\label{fig:gpt4-prompt-colors}
\end{figure}
\clearpage
\subsubsection{Position}
\begin{figure}[H]
\centering
\fbox{%
\begin{minipage}{1.0\textwidth}
\ttfamily
You are a multimodal large-language model tasked with evaluating images generated by a text-to-image model. Your goal is to assess each generated image based on specific aspects and provide a detailed critique, along with a scoring system. The final output should be formatted as a JSON object containing individual scores for each aspect and an overall score. The keys in the JSON object should be: \texttt{position\_accuracy}, \texttt{occlusion\_management}, \texttt{perspective\_consistency}, and \texttt{overall\_score}. Below is a comprehensive guide to follow in your evaluation process: Your evaluation should focus on these aspects:

\textbf{1. Key Evaluation Aspects and Scoring Criteria}:
For each aspect, provide a score from 0 to 10, where 0 represents poor performance and 10 represents excellent performance. For each score, include a short explanation or justification (1-2 sentences) explaining why that score was given. The aspects to evaluate are as follows: 

\textbf{a) Positional Accuracy}:
Assess the matching accuracy between spatial position and prompt description.
Score: 0 (totally wrong) to 10 (postion correct)

\textbf{b) Occlusion Management}:
Evaluate position discernibility in the presence of occlusion.
Score: 0 (fully occlusion) to 10 (clearly dsiplay the relationship).

\textbf{c) Perspective Consistency}:
Assess the rationality of perspective relationship and spatial depth.
Score: 0 (perspective contradiction) to 10 (completely reasonable).

\textbf{2. Overall Score}: 
After scoring each aspect individually, provide an overall score, representing the model's general performance on this image. This should be a weighted average based on the importance of each aspect to the prompt or an average of all aspects.
\end{minipage}
}
\caption{Verifier prompt for images for positions.}
\label{fig:gpt4-prompt-position}
\end{figure}
\clearpage

\subsubsection{Color Attribution}
\begin{figure}[H]
\centering
\fbox{%
\begin{minipage}{1.0\textwidth}
\ttfamily
You are a multimodal large-language model tasked with evaluating images generated by a text-to-image model. Your goal is to assess each generated image based on specific aspects and provide a detailed critique, along with a scoring system. The final output should be formatted as a JSON object containing individual scores for each aspect and an overall score. The keys in the JSON object should be: \texttt{attribute\_binding}, \texttt{contrast\_effectiveness}, \texttt{material\_consistency}, and \texttt{overall\_score}. Below is a comprehensive guide to follow in your evaluation process: Your evaluation should focus on these aspects:

\textbf{1. Key Evaluation Aspects and Scoring Criteria}:
For each aspect, provide a score from 0 to 10, where 0 represents poor performance and 10 represents excellent performance. For each score, include a short explanation or justification (1-2 sentences) explaining why that score was given. The aspects to evaluate are as follows: 

\textbf{a) Attrribute Binding}:
Correct binding of colors to designated objects (no color mismatches).
Score: 0 (color mismatch) to 10 (correct binding).

\textbf{b) Contrast Effectiveness}:
Evaluate the difference between foreground and background colors.
Score: 0 (similar colors, difficult to distinguish) to 10 (high contrast).

\textbf{c) Material Consistency}:
Assess the coordination of color and material performance.
Score: 0 (material conflicts) to 10 (perfect harmony).

\textbf{2. Overall Score}: 
After scoring each aspect individually, provide an overall score, representing the model's general performance on this image. This should be a weighted average based on the importance of each aspect to the prompt or an average of all aspects.
\end{minipage}
}
\caption{Verifier prompt for images for color attribution.}
\label{fig:gpt4-prompt-color-attribution}
\end{figure}

\subsection{Reflection Prompt}
\begin{figure}[H]
\centering
\fbox{%
\begin{minipage}{1.0\textwidth}
\ttfamily
You are an expert assistant for generating image improvement instructions. Analyze the original prompt, the updated prompt to generate the image, the evaluation of the generated image, and the generated image, give instructions to create specific technical directions following these guidelines:

\textbf{1. Structure and Focus Areas}:
Focus strictly on this aspect: Prompt Following.

\textbf{2. Detailed Requirements for Each Aspect}:
A. Prompt Following Instructions: Examine the original prompt sentence by sentence. List exact discrepancies between the bad image and prompt specifications. Use direct action verbs: Add, Remove, Replace, Reposition, Adjust, to modify the image. Specify precise locations and modification commands. Never use vague terms like ensure or confirm.

\textbf{3. Format Specifications}:
Use exact section headers without markdown:1. Prompt Following:\textbackslash n-\textbackslash n 
Each instruction must start with a hyphen and complete command. Include spatial references and implementation details. Omit sections with no required improvements. Never include explanations or examples.

\textbf{4. Content Principles}:
Every instruction must be directly executable by an artist. Prioritize critical errors first. Describe only missing or incorrect elements. Use imperative verb forms exclusively. Maintain technical specificity without assumptions.
\end{minipage}
}
\caption{Prompt for generating reflection instructions.}
\label{fig:gpt4-prompt-reflection}
\end{figure}

\subsection{Refine Prompt}
\begin{figure}[H]
\centering
\fbox{%
\begin{minipage}{1.0\textwidth}
\ttfamily
You are a multimodal large-language model tasked with refining user's input prompt to 
create images using a text-to-image model. Given a original prompt, a current prompt,
a batch of images generated by the prompt, a reflection prompt about the generated images and their corresponding assessments 
evaluated by a multi-domain scoring system, your goal is to refine the current prompt to 
improve the overall quality of the generated images. You should analyze the strengths
and drawbacks of current prompt based on the given images and their evaluations. Consider
aspects like subject, scene, style, lighting, tone, mood, camera style, composition, and
others to refine the current prompt. Do not alter the original description from
the original prompt. The refined prompt should not contradict with the reflection prompt. Directly output the refined prompt without any other text.

\par\vspace{1em}

Some further instructions you should keep in mind:

1) The current prompt is an iterative refinement of the original prompt.

2) In case the original prompt and current prompt are the same, ignore the current prompt.

3) In some cases, some of the above-mentioned inputs may not be available. For example, the images,
the assessments, etc. In such situations, you should still do your best, analyze the inputs carefully,
and arrive at a refined prompt that would potentially lead to improvements in the final generated images.

4) When the evaluations are provided, please consider all aspects of the evaluations very carefully.
\end{minipage}
}
\caption{Prompt for refinement.}
\label{fig:gpt4-prompt-refinement}
\end{figure}
\clearpage

\subsection{Chain-of-Though Image Reflection Prompt}
\begin{figure}[H]
\centering
\fbox{%
\begin{minipage}{1.0\textwidth}
\ttfamily
You are a multimodal analysis assistant. Given a prompt and two generated images (left and right), your task is to analyze and compare both images with respect to their alignment to the given prompt, decide which image better matches the prompt, then generate concise editing instructions to modify the inferior image to become the superior image. Follow the step-by-step instructions below:

\par\vspace{1em}

1. Analyze both images carefully and identify key differences regarding: missing elements, incorrect object attributes (color, size, position, number, etc.), incorrect spatial or logical relationships between objects, presence of unnecessary elements, etc.

2. Based on the analysis, determine which image better aligns with the prompt. Output "left" if the left image is better; output "right" if the right image is better.

3. Generate exactly one most important editing instruction that will modify the inferior image to closely match the better image. Follow these guidelines:

\setlength{\parindent}{2em} 
    (1) Use concise, accurate, actionable imperative sentences.
    
    (2) **DO NOT explicitly mention specific images in your response, like "the left image", "the right image", or similar words!**
    
    (3) Avoid vague or redundant instructions, such as "ensure" or "verify".
    
    (4) Example instructions:

\setlength{\parindent}{4em}
        - "Add a dirt road in the foreground extending into the background."
        
        - "Remove a cluster of white, fluffy cotton grass plants in the foreground on the rocky shore."
        
        - "Swap the vampire with a woman with long, wavy blonde hair."
        
        - "Make the image look like it's from an ancient Egyptian mural."
        
        - "Turn the color of golden shield to gray."

\par\vspace{1em}

\setlength{\parindent}{0em}
Format your final response strictly in JSON format:

```json

\{

\setlength{\parindent}{2em}
    "Analysis": "<detailed analysis of key differences between the two images in relation to the prompt>",
    
    "Result": "<left/right>",
    
    "Instructions": "<instruction>"
\setlength{\parindent}{0em}

\}

```

\end{minipage}
}
\caption{Prompt for generating chain-of-thought image reflection annotations.}
\label{fig:cot-prompt-reflection}
\end{figure}

\clearpage